\pdfoutput=1

\documentclass[11pt]{article}

\usepackage[preprint]{acl}

\usepackage{inconsolata}
\usepackage{xcolor}
\usepackage{times}
 
\usepackage{booktabs}
\usepackage{multirow}
\usepackage{tablefootnote}
\usepackage{array}
\usepackage{pdfpages}
\usepackage{colortbl}
\usepackage{amssymb}  

\usepackage{amsmath}
\usepackage{amsfonts}

\usepackage{graphicx}
\usepackage{float}
\usepackage{subcaption}
\usepackage{paralist}
\usepackage{enumitem}
\usepackage{titlesec}

\usepackage[T1]{fontenc}

\usepackage[utf8]{inputenc}

\usepackage{microtype}

\usepackage{inconsolata}

\usepackage{graphicx}

\title{MINTQA: A Multi-Hop Question Answering Benchmark for Evaluating LLMs on New and Tail Knowledge}

\author{
    Jie He$^1$\footnotemark[1] \quad 
    Nan Hu$^2$\thanks{\ \ Equal Contribution.} \quad 
    Wanqiu Long$^1$\footnotemark[1] \quad 
    Jiaoyan Chen$^3$ \quad 
    \textbf{Jeff Z. Pan}$^1$ \; \\
     $^1$ School of Informatics, University of Edinburgh, UK  \\
    $^2$ Southeast University, Nanjing, Jiangsu, China \\
    $^3$ University of Manchester, UK\\
    \normalsize{\texttt{ j.he@ed.ac.uk, nanhu@seu.edu.cn, }} \normalsize{\texttt{j.z.pan@ed.ac.uk}} \\
}

\begin{document}
\maketitle
\begin{abstract}
\textbf{L}arge \textbf{l}anguage \textbf{m}odels (LLMs) have demonstrated impressive capabilities in various reasoning tasks but face significant challenges with complex, knowledge-intensive multi-hop queries, particularly those involving new or long-tail knowledge. Existing benchmarks often fail to fully address these challenges. To bridge this gap, we introduce \textbf{MINTQA} (\textbf{M}ult\textbf{i}-hop \textbf{Q}uestion \textbf{A}nswering on \textbf{N}ew and \textbf{T}ail Knowledge), a comprehensive benchmark to evaluate LLMs' capabilities in multi-hop reasoning across four critical dimensions: question handling strategy, sub-question generation, retrieval-augmented generation, and iterative or dynamic decomposition and retrieval. MINTQA comprises 10,479 question-answer pairs for evaluating new knowledge and 17,887 pairs for assessing long-tail knowledge, with each question equipped with corresponding sub-questions and answers. Our systematic evaluation of 22 state-of-the-art LLMs on MINTQA reveals significant limitations in their ability to handle complex knowledge base queries, particularly in handling new or unpopular knowledge. Our findings highlight critical challenges and offer insights for  advancing multi-hop reasoning capabilities\footnote{The MINTQA benchmark is available at \href{ https://github.com/probe2/multi-hop/}{ https://github.com/probe2/multi-hop/.}}.

\end{abstract}

\section{Introduction}

\textbf{L}arge \textbf{l}anguage \textbf{m}odels (LLMs) have demonstrated remarkable capabilities in question answering tasks \cite{Kamalloo2023EvaluatingOQ,Wang2024NoNF}. However, they face significant challenges when handling multi-hop queries requiring specific knowledge or recent information. While \textbf{R}etrieval-\textbf{A}ugmented \textbf{G}eneration (RAG) offers an effective strategy by incorporating external knowledge during response generation \cite{10.1145/3673791.3698415,Islam2024OpenRAGER}, its effectiveness in multi-hop reasoning scenarios presents unique challenges.

\begin{figure}
    \centering
    \includegraphics[width=0.9\linewidth]{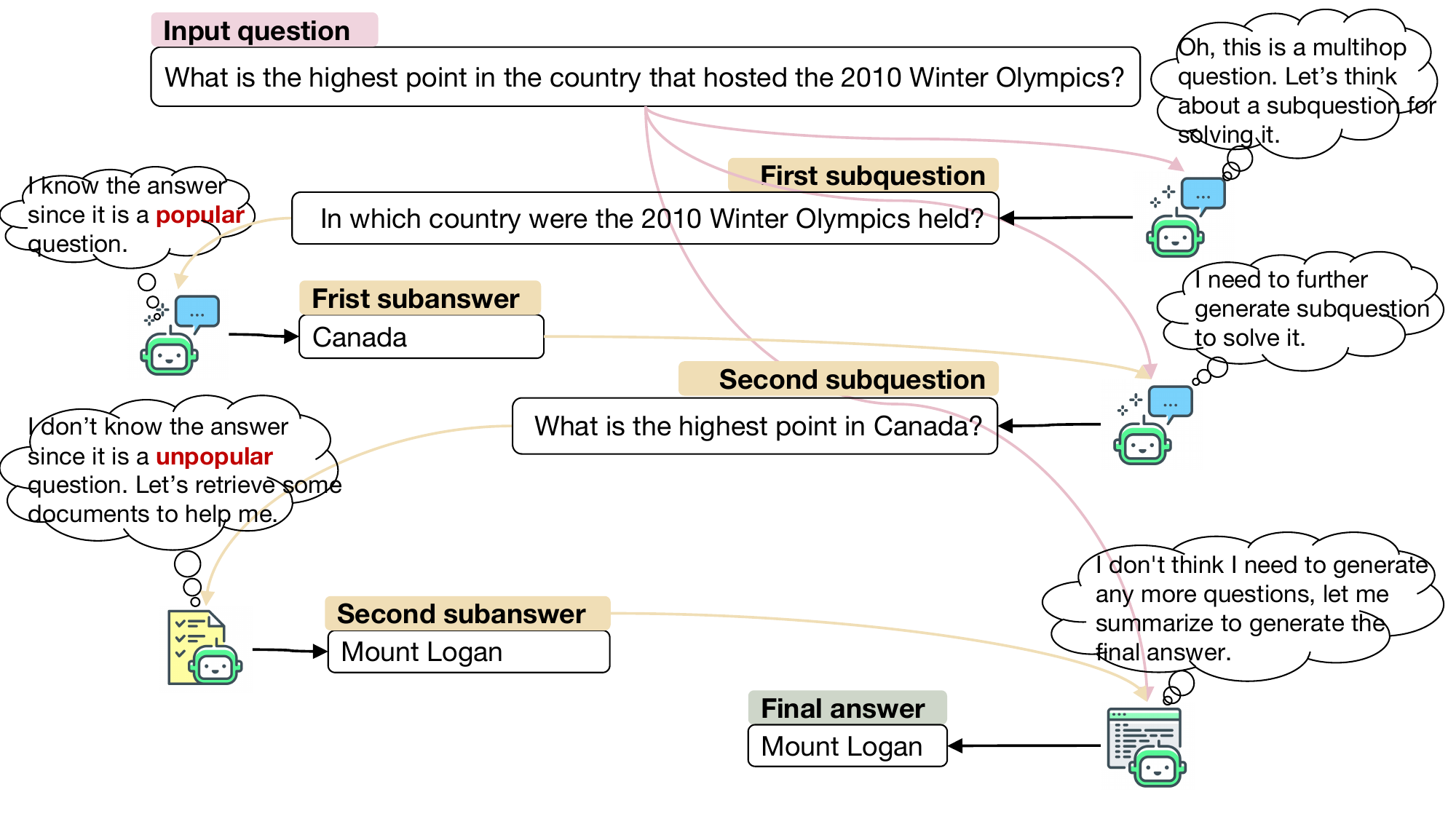}
    \caption{\textbf{A example for our benchmark:}  Given a complex question, the model must decide whether to decompose it into sub-questions and determine if external knowledge retrieval is required. }
    \label{fig:intro}
    \vspace{-0.4cm}
\end{figure}

Consider a complex question: ``What is the highest point in the country that hosted the 2010 Winter Olympics?'' As illustrated in Figure~\ref{fig:intro}, to answer such questions, models need to decompose the question into sub-questions (e.g., ``In which country were the 2010 Winter Olympics held?'' followed by "What is the highest point in Canada?''). For each sub-question, models must decide whether to use parametric knowledge or perform retrieval. For instance, Olympic host countries might be reliably answered using parametric knowledge, while specific geographical details like the highest point may require retrieval. This process becomes particularly challenging when questions involve new or unpopular knowledge, requiring models to effectively coordinate between knowledge source selection, question decomposition, and multi-step reasoning.


Current frameworks for evaluating LLMs on \textbf{q}uestion \textbf{a}nswering (QA) have several critical limitations. First, studies such as \cite{Sun2023HeadtoTailHK,Maekawa2024RetrievalHO,Zhang2024RetrievalQAAA} focus primarily on single-hop queries, leaving complex multi-hop questions largely unexplored. Second, while multi-hop benchmarks such as MultiHop-RAG \cite{Tang2024MultiHopRAGBR} assess retrieval effectiveness, they overlook the crucial decision-making process of when and how to retrieve and fail to systematically evaluate the interaction between question decomposition and retrieval, a capability essential for real-world applications. Furthermore, existing works like FanoutQA \cite{Zhu2024FanOutQAAM} and HotpotQA \cite{Yang2018HotpotQAAD} lack an assessment of how models handle queries containing new or unpopular knowledge, which presents unique challenges in both decomposition and retrieval.

To bridge these gaps, we propose MINTQA, a benchmark for evaluating LLMs on complex multi-hop questions across two critical dimensions: \textbf{Unpopular knowledge} (information appearing infrequently in training corpora) and \textbf{New Knowledge} (recently emerged entities or relationships). We construct MINTQA by systematically collecting knowledge triplets from the English Wikidata and using GPT-4o to generate multi-hop questions spanning one to four hops. The benchmark comprises two sub-datasets: \textbf{\textsc{MINTQA-pop}} (17,887 examples) focusing on unpopular/popular knowledge, and \textbf{\textsc{MINTQA-ti}} (10,479 examples) examining new/old knowledge, with each example including sub-questions and answers for fine-grained analysis of models' reasoning processes. 

Our framework evaluates LLMs across five critical aspects: 1) Evaluating LLMs using their parametric knowledge (Section \ref{sec5}); 2) Question handling strategies selection (Section \ref{sec6}); 3) Retrieval-augmented generation (Section \ref{sec7}); 4) Sub-question Generation (Section \ref{sec8}); 5) Iterative or dynamic decomposition and retrieval (Section \ref{sec9}).

The comprehensive evaluation of 22 state-of-the-art LLMs reveals several key findings. First, performance varies in \textsc{MINTQA-pop} and \textsc{MINTQA-ti}, and strategies like retrieval and question decomposition show varying effectiveness on them. Second, larger models generally demonstrate better awareness of their knowledge boundaries, particularly for \textsc{MINTQA-ti} questions involving new information, but they can be overconfident in some cases, while smaller models often fail to assess question complexity, answering directly instead of selecting appropriate strategies. Third, performance consistently declines and the effectiveness of retrieval decreases with increasing reasoning hops. 

We also implement the dynamic retrieval method \cite{Ni2024WhenDL} on our benchmark, which relies on model-based decisions to optimize retrieval frequency. However, maintaining performance while reducing retrieval frequency remains challenging on MINTQA, and some models show excessive retrieval dependency. Additionally, the best-performing model, LLaMA 3.1-70B, only achieves an overall accuracy of 62.33\% on MINTQA, highlighting the significant challenges in complex multi-hop reasoning even with retrieval.

The key contributions of this study are:
\begin{compactenum}
    \item We introduce MINTQA, a novel benchmark for evaluating LLMs' multi-hop reasoning capabilities across different knowledge types, with reasoning chains of varying complexity.
    \item We present a systematic evaluation framework examining key aspects of multi-hop QA, enabling comprehensive analysis of models' reasoning capabilities and the effectiveness of strategies for enhancing LLM performance.
   \item Our evaluation on 22 state-of-the-art LLMs reveals their limitations in complex multi-hop reasoning, offering valuable insights to enhance their capabilities in multi-hop QA.
\end{compactenum}

\section{Related Work}
\subsection{Multi-hop Question Answering (QA)}
Multi-hop QA challenges LLMs by requiring synthesis and reasoning across multiple sources \cite{ feng2020learningrecoverreasoningchains, Khashabi2019OnTP, huang-chang-2023-towards,xiong-etal-2025-deliberate}. While researchers have proposed decomposing complex questions into sub-questions \cite{Min2019CompositionalQD, Wang2022LocateTA, Wang2023SelfpromptedCO, Liu2024RAISFLT}, generating relevant sub-questions and reasoning chains remains challenging. Existing benchmarks \cite{Zhang2024RetrievalQAAA, Zhu2024FanOutQAAM, Tang2024MultiHopRAGBR} assess retrieval and multi-hop reasoning, but overlook when and how to retrieve, interactions between decomposition and retrieval, or queries with new and unpopular knowledge. Our MINTQA fills these gaps by systematically evaluating LLMs’ on multi-hop QA.

\begin{figure*}[!th]
    \centering
    \includegraphics[width=0.65\linewidth]{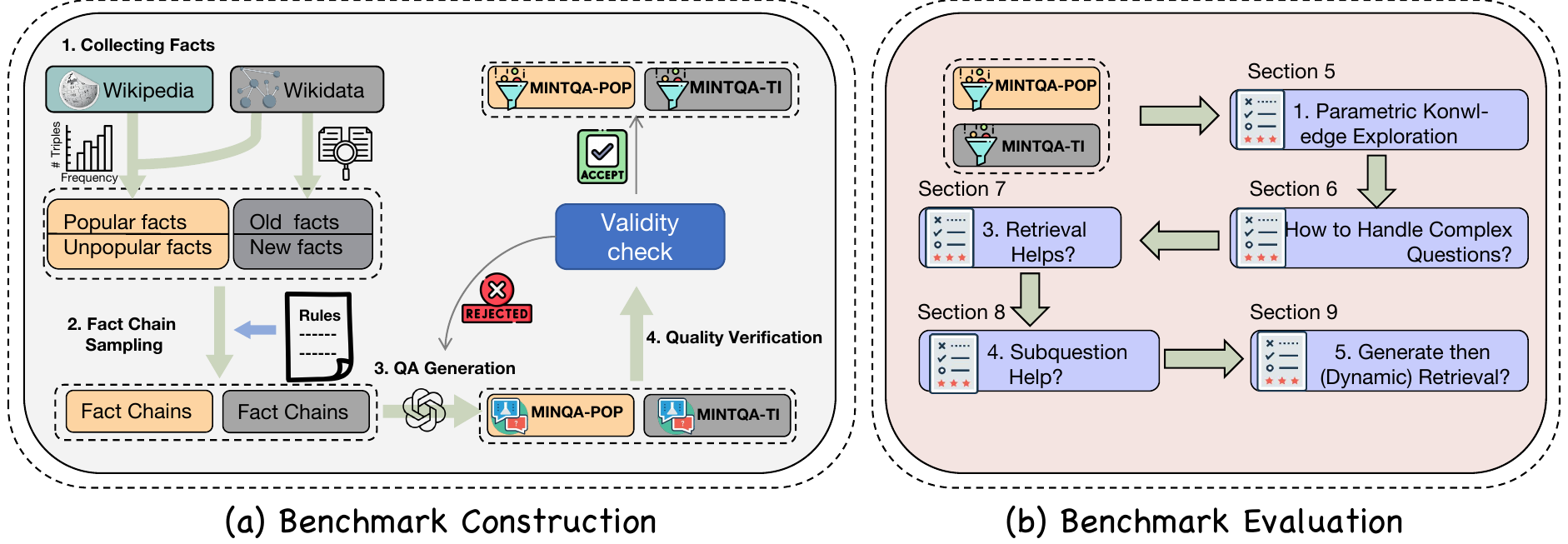}
\caption{Two components of our work: (a) we sample different types of facts from Wikidata to generate complex questions; (b) we conduct a comprehensive evaluation of existing LLMs from five perspectives.}
    \label{fig:main_figure}
    \vspace{-0.3cm}
\end{figure*}
\subsection{Retrieval Augmented Generation (RAG)}
RAG enhances LLMs' performance in multi-question answering by providing access to external documents \cite{Lewis2020RetrievalAugmentedGF, Xiong2020AnsweringCO}, particularly for knowledge-intensive tasks \cite{Yu2020ASO, Zhu2023LargeLM}. In sub-question generation, RAG can verify and correct LLMs' outputs \cite{Zhao2023VerifyandEditAK,Shi2024RetrievalenhancedKE}. However, irrelevant retrievals can introduce noise, and external knowledge may override model's inherent knowledge \cite{Xu2023SearchintheChainTA, Li2022LargeLM}, while adding computational overhead \cite{Zhu2024AcceleratingIO}. While \citet{Jeong2024AdaptiveRAGLT} propose using a classifier to determine retrieval necessity, our research investigates LLMs' inherent ability to recognize when retrieval is needed for sub-questions.

\subsection{Evaluation of LLMs}
Existing QA datasets for evaluating retrieval-augmented LLMs fall into two categories: (1) Reasoning-focused datasets \cite{Ho2020ConstructingAM,Yang2018HotpotQAAD,Sen2022MintakaAC}, such as MuSiQue \cite{Trivedi2021MM}, FanOutQA \cite{Zhu2024FanOutQAAM}, and MultiHop-RAG \cite{Tang2024MultiHopRAGBR} that emphasize multi-hop reasoning across multiple documents; (2) Long-tail question datasets \cite{Mallen2022WhenNT, Zhang2024RetrievalQAAA}, including WiTQA \cite{Maekawa2024RetrievalHO} focusing on rare single-hop queries and Head-to-Tail \cite{Sun2023HeadtoTailHK} examining entity and relationship popularity to highlight the value of knowledge graphs. Our work extends these by evaluating both long-tail and new-fact multi-hop QA, while analyzing models' sub-question generation and retrieval capabilities.

\section{Benchmark Construction}
\label{sec3}
This section presents our comprehensive methodology for constructing two multi-hop QA benchmarks: \textsc{MINTQA-pop} and \textsc{MINTQA-ti}, designed to evaluate LLM across two critical dimensions: knowledge popularity (popular versus unpopular) and temporal knowledge (new versus old). We first present the data construction methodology for \textsc{MINTQA-pop} (Section \ref{sec3.1}). We then detail the construction process of \textsc{MINTQA-ti}, which follows a similar procedure but focuses on new/old knowledge (Section \ref{sec:MINTQA-ti_construction}). Finally, we describe our QA generation process (Section \ref{sec3.3}) and present comprehensive statistics of our constructed datasets (Section \ref{sec3.4}). Figure \ref{fig:main_figure} outlines our benchmark construction and evaluation process.


\subsection{Data Construction of \textsc{MINTQA-pop}} 
\label{sec3.1}
\label{sec:MINTQA-pop_construction}
\textbf{Collecting Facts} We gather a collection of facts with popularity, denoted as $\mathcal{G}_{pop} = \{(s, r, o), p|(s, r, o) \in \mathcal{G}, p \in \mathbb{Z}^+)\} $, where $\mathcal{G}$ refers to Wikidata, \((s, r, o)\) represents a triple, \( p \) indicates the popularity as the positive integer set $\mathbb{Z}^+$. The triples are extracted from Wikipedia (version 2024-05-01). Specifically, we extract raw triples in the format of (Head Span, Relation, Tail Span) from Wikipedia passages using an existing information extraction tool\footnote{https://github.com/Babelscape/rebel}. These raw triples are linked to Wikidata (version 2024-04-22) using WikiMapper\footnote{https://github.com/jcklie/wikimapper}, producing structured triples with Wikidata IDs \((s, r, o)\). We only keep the triples \((s, r, o)\) existing in Wikidata. The popularity \( p \) of each triples is calculated as the frequency of its occurrence across the entire Wikipedia corpus.


\noindent \textbf{Sampling fact chains} We sample facts from $\mathcal{G}_{pop}$ and concatenate them into a chain $\mathcal{FC} = \{(s_1, r_1, o_1), \ldots, (s_n, r_n, o_n)\}$ as the grounded facts of a multi-hop question. We categorize facts in $\mathcal{G}_{pop}$ based on their popularity scores into two distinct sets: \textbf{unpopular} knowledge ($\mathcal{P}_{\text{unpop}} = [1,10)$) and \textbf{popular} knowledge ($\mathcal{P}_{\text{pop}} = [50,\infty)$). A fact chain $\mathcal{FC}$ is constructed as an ordered sequence of connected triples: $\mathcal{FC} = \{(s_1,r_1,o_1),\ldots,(s_n,r_n,o_n)\}$, where $n \leq 4$ and each triple can be either popular or unpopular. This construction follows four key constraints:
\begin{compactenum}
    \item Connectivity: $o_i = s_{i+1}$ for all $i \in \{1,\ldots,n-1\}$.
    \item Acyclicity: $o_i \neq s_j$ for all $i, j \in \{1,\ldots,n\}$.
    \item Uniqueness: No fact chain $\mathcal{FC}$ can be a sub-chain of another fact chain.
    \item No Shortcuts: For each fact chain $\mathcal{FC}$, there does not exist a triple $(s_i,r,o_j)$ in $\mathcal{G}_{pop}$ such that $j > i+1$, where $i \in \{1,\ldots,n-1\}$ and $j \in \{2,\ldots,n\}$.
\end{compactenum}

\subsection{Data Construction of \textsc{MINTQA-ti}}
\label{sec:MINTQA-ti_construction}
Building on the methodology established for \textsc{MINTQA-pop}, we construct \textsc{MINTQA-ti}, focusing on old and new knowledge. To construct the dataset, we extract two versions of Wikidata: 2021-06-21 and 2024-06-05. We identify triples that are either common to both versions or differ between them. These triples form the knowledge graph $\mathcal{G}_{ti}$. We define \textbf{old} knowledge as triples present in both Wikidata versions, and \textbf{new} knowledge as triples only appearing in the newer version, characterized by a new subject, relation, or object. Following the same chain construction principles outlined in Section \ref{sec:MINTQA-pop_construction}, we create fact chains combining new and old knowledge from $\mathcal{G}_{ti}$.


\subsection{QA Generation and Verification} 
\label{sec3.3}
Following WiTQA \cite{Maekawa2024RetrievalHO}, we employ GPT-4o to automatically generate questions from extracted triplets, overcoming the diversity and scalability issues of template-based methods like PopQA \cite{Mallen2022WhenNT} and the high costs of manual annotation. Given a fact chain $\mathcal{FC} = \{(s_1, r_1, o_1), \ldots, (s_n, r_n, o_n)\}$, where $o_{i-1} = s_n, i \in \{1,\ldots,n\}$, we aim to generate a question about $s_1$ that yields $o_n$ as the answer. To enhance generation quality, we provide one demonstration example per hop. And to ensure validity, we verify questions by having the model answer them using source contexts; only questions yielding $o_n$ are retained. Invalid questions are regenerated up to three times, and unsatisfactory examples are discarded. For multi-hop questions (hop count $\geq 2$), sub-questions for each intermediate fact are also generated and validated. Examples are included in the dataset only if the main question and all sub-questions pass validation. This process filtered out 138 and 67 examples from \textsc{MINTQA-pop} and \textsc{MINTQA-ti}, respectively. Prompts and examples are in Appendices \ref{data_creation} and \ref{case_study}.


\begin{table}[!t]
\tiny
    \centering
    \begin{tabular}{c|cccc|c}
\hline 
\textbf{\textsc{MINTQA-pop}} & \textbf{1-hop} & \textbf{2-hop} & \textbf{3-hop} & \textbf{4-hop} & \textbf{Total} \\
\hline
\#Samples & 5,894 & 4,428 & 4,664 & 2,901 & 17,887 \\
\hline
\#Input Tok. & 52,488 & 59,699 & 74,363 & 57,579 & 244,129 \\
\#Input Vocab & 7,468 & 5,410 & 7,094 & 4,491 & 18,852 \\
Avg. In. Len. & 8.91 & 13.48 & 15.94 & 19.85 & 13.65 \\
\#Output Tok. & 11,526 & 6,249 & 6,196 & 5,610 & 29,581 \\
\#Output Vocab & 3,398 & 420 & 228 & 210 & 3,721 \\
Avg. Out. Len. & 1.96 & 1.41 & 1.33 & 1.93 & 1.65 \\
Avg. Ctx. Len. & 32.93 & 375.93 & 549.18 & 706.12 & 361.63 \\
\#Relations & 124 & 84 & 85 & 98 & 140 \\
\#Entities & 7,482 & 4,357 & 5,191 & 3,180 & 18,501 \\
\hline 
\textbf{\textsc{MINTQA-ti}} &  &  &  &  &  \\
\hline
\#Samples & 3,949 & 2,198 & 2,057 & 2,275 & 10,479 \\
\#Input Tok. & 34,014 & 30,471 & 34,106 & 44,959 & 143,550 \\
\#Input Vocab & 5,586 & 3,113 & 2,439 & 2,695 & 10,345 \\
Avg. In. Len. & 8.61 & 13.86 & 16.58 & 19.76 & 13.70 \\
\#Output Tok. & 8,064 & 5,738 & 4,365 & 5,086 & 23,253 \\
\#Output Vocab & 1,710 & 1,004 & 996 & 809 & 3,318 \\
Avg. Out. Len. & 2.04 & 2.61 & 2.12 & 2.24 & 2.22 \\
Avg. Ctx. Len. & 56.44 & 321.32 & 487.22 & 645.61 & 324.47 \\
\#Relations & 123 & 147 & 149 & 147 & 189 \\
\#Entities & 4,096 & 2,484 & 2,250 & 2,346 & 9,616 \\
\hline 
    \end{tabular}
    \caption{Data statistics of MINTQA.}
    \label{tab:data_statis_table}
    \vspace{-0.3cm}
\end{table}

\subsection{Dataset Statistics} 
\label{sec3.4}
Table \ref{tab:data_statis_table} summarizes the statistics of the \textsc{MINTQA-pop} and \textsc{MINTQA-ti} datasets, which exhibit diverse coverage across multiple dimensions. \textsc{MINTQA-pop} contains 17,887 examples and \textsc{MINTQA-ti} 10,479, with over 2,000 examples per hop category, ensuring robust evaluation. The datasets include 18,501 and 9,616 entities, and 140 and 198 relationships, respectively, demonstrating their diversity. As the number of hops increases, the average context length grows, requiring models to retrieve more documents and face greater challenges. For details, see the App. \ref{data_analy_app}.

\section{Experimental Setup}
\subsection{Language Models and Configurations}

\textbf{Large Language Models} We evaluate state-of-the-art LLMs across various architectures and model sizes: GPT-3.5, GPT-4o, GPT-4o mini, LLaMA-3.1/3.2 \cite{grattafiori2024llama3herdmodels}, Gemma-2 \cite{gemmateam2024gemma2improvingopen}, Mistral \cite{jiang2023mistral7b}, Phi-3 \cite{abdin2024phi3technicalreporthighly}, and Qwen2.5 \cite{hui2024qwen25codertechnicalreport}. All models are instruct versions. For simplicity, we omitted the ``instruct'' name in the result presentation. To ensure reproducibility, we set the temperature parameter to 0 across all models and accelerated inference using vLLM \cite{Kwon2023EfficientMM}. For more details, please refer to Appendix~\ref{appdix_exp}.

\subsection{Evaluation metrics}

We adopt \textit{Accuracy} (Acc) as our evaluation metric across all experiments by determining whether the ground-truth answer is present in the model's predicted text across all experiments, consistent with established benchmarks in factual knowledge assessment \cite{Ren2023InvestigatingTF,Maekawa2024RetrievalHO, Mallen2022WhenNT}.

\begin{figure}[!th]
    \centering
    \includegraphics[width=0.6\linewidth]{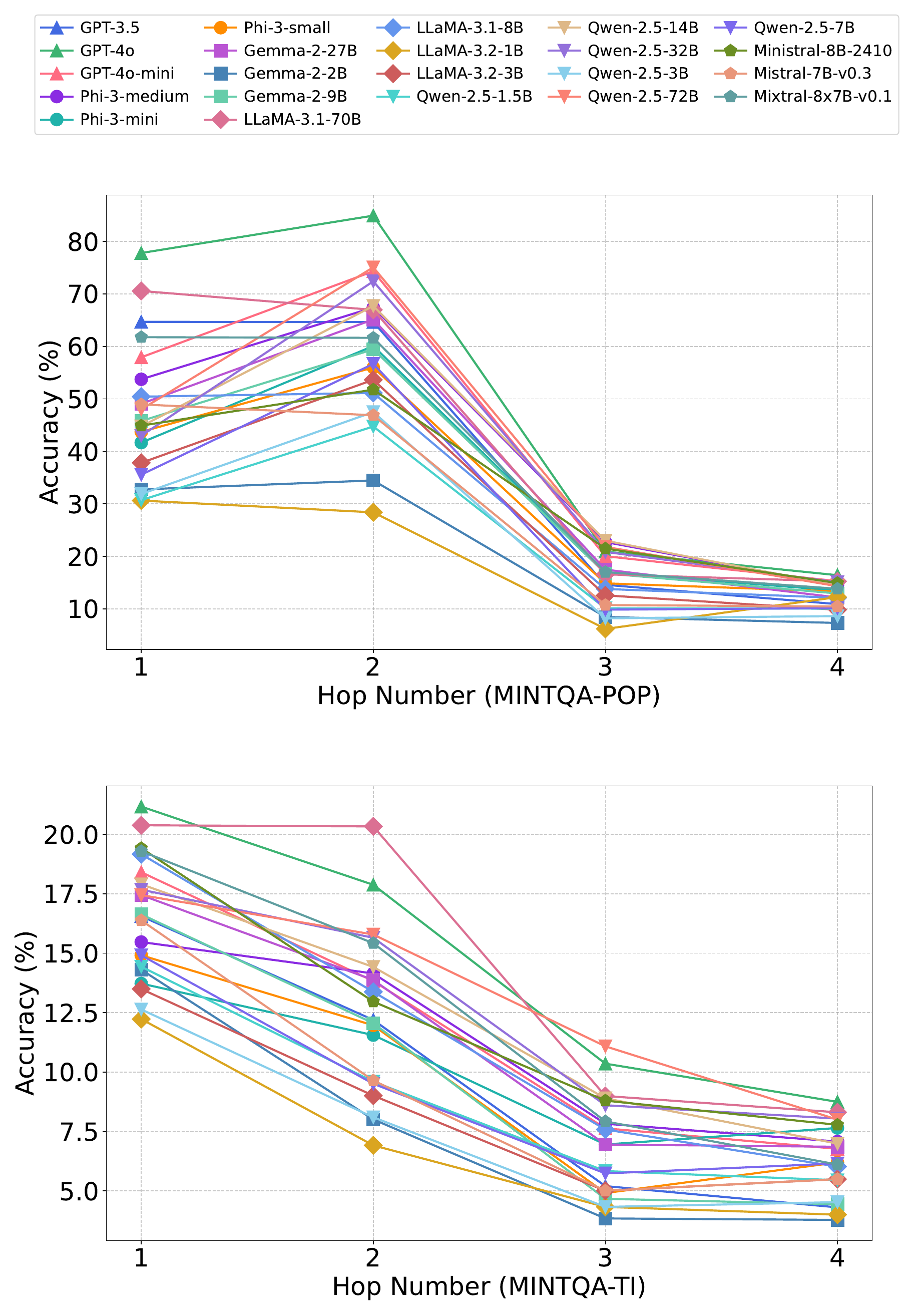}
    \caption{Zero-shot accuracy of different LLMs across various hops.}
    \label{fig:zero-shot}
    \vspace{-0.1cm}
\end{figure}
\section{LLMs' Performance on MINTQA with Parametric Knowledge}
\label{sec5}
We evaluate LLMs on MINTQA using their parametric knowledge to understand intrinsic model capabilities and dataset challenges. The results are shown in Figure \ref{fig:zero-shot} and further elaborated in App. \ref{app_zero_shot}. 
Our findings reveal significant performance gaps between the \textsc{MINTQA-pop} and \textsc{MINTQA-ti} . Models perform reasonably on \textsc{MINTQA-pop} (e.g., GPT-4o: 77.79\%, LLaMA3.1-8B: 50.42\% for single-hop questions) but struggle on \textsc{MINTQA-ti}, with GPT-4o's accuracy dropping to 21.17\% for single-hop questions. This confirms \textsc{MINTQA-ti}’s effectiveness in evaluating knowledge beyond training data, and low performance across models from LLaMA-3.2-1B (7.78\%) to GPT-4o (21.17\%) demonstrates scaling model size alone doesn’t address this. Moreover, increased reasoning complexity further highlights these limitations. On \textsc{MINTQA-pop}, performance drops sharply for three-hop (20.03\%) and four-hop (16.41\%) questions, while on \textsc{MINTQA-ti}, accuracy consistently declines with complexity.

\section{Evaluating LLMs' Decision-Making Capabilities in Multi-hop QA}
\label{sec6}


\begin{table}[!t]
\tiny
\centering
\begin{tabular}{l|cc|cc}
\hline
\multirow{2}{*}{Model} & \multicolumn{2}{c|}{\textsc{MINTQA-pop}} & \multicolumn{2}{c}{\textsc{MINTQA-ti}} \\
\cline{2-5}
& Acc & F1 & Acc & F1 \\
\hline
GPT-3.5 & 54.82 & 41.21 & 52.04 & 26.67 \\
GPT-4o-mini & 65.61 & 48.67 & 59.56 & 28.72 \\
GPT-4o & 68.84 & 51.35 & 65.46 & 38.18 \\
\hline
LLaMA-3.1-8B & 59.93 & 25.83 & 55.55 & 26.80 \\
LLaMA-3.1-70B & 70.03 & 44.88 & 63.26 & 29.54 \\
LLaMA-3.2-1B & 67.05 & 26.76 & 62.32 & 25.59 \\
LLaMA-3.2-3B & 56.40 & 32.91 & 47.37 & 33.65 \\
\hline
Qwen2.5-1.5B & 64.01 & 26.47 & 58.42 & 24.92 \\
Qwen2.5-3B & 47.08 & 38.89 & 48.09 & 41.86 \\
Qwen2.5-7B & 70.90 & 48.23 & 62.22 & 34.47 \\
Qwen2.5-14B & 69.31 & 42.08 & 62.39 & 26.28 \\
Qwen2.5-32B & 28.74 & 29.80 & 18.69 & 11.04 \\
Qwen2.5-72B & 65.39 & 55.33 & 51.11 & 37.00 \\
\hline
Gemma-2-2B & 5.39 & 3.84 & 21.07 & 14.53 \\
Gemma-2-9B & 70.90 & 46.18 & 63.05 & 30.39 \\
Gemma-2-27B & 73.22 & 55.67 & 63.64 & 36.15 \\
\hline
Phi-3-mini & 64.47 & 26.44 & 56.60 & 24.72 \\
Phi-3-small & 75.79 & 55.56 & 62.53 & 40.50 \\
Phi-3-medium & 28.43 & 21.69 & 18.50 & 10.64 \\
\hline
Mistral-7B-v0.3 & 37.58 & 30.94 & 26.28 & 18.88 \\
Mixtral-8x7B-v0.1 & 28.00 & 15.14 & 18.48 & 10.51 \\
Ministral-8B-2401 & 67.07 & 27.35 & 62.27 & 25.66 \\
\hline
\end{tabular}
\caption{The model's accuracy and F1 score for the task of determining question retrieval, sub-question generation, or direct answering.}
\label{tab:ex2-part1}
\vspace{-0.4cm}
\end{table}
\begin{table}[!t]
\tiny
\centering
\begin{tabular}{l|cc|cc}
\hline
\multirow{2}{*}{Model} & \multicolumn{2}{c|}{\textsc{MINTQA-pop}} & \multicolumn{2}{c}{\textsc{MINTQA-ti}} \\
\cline{2-5}
& Acc & F1 & Acc & F1 \\
\hline
GPT-3.5 & 49.32 & 49.15 & 47.32 & 47.27 \\
GPT-4o-mini & 47.77 & 46.98 & 47.48 & 47.14 \\
GPT-4o & 37.13 & 33.19 & 44.67 & 43.83 \\
\hline
LLaMA-3.1-8B & 30.88 & 23.66 & 43.05 & 30.15 \\
LLaMA-3.1-70B & 54.60 & 54.53 & 48.46 & 47.70 \\
LLaMA-3.2-1B & 31.40 & 24.71 & 43.07 & 30.74 \\
LLaMA-3.2-3B & 30.82 & 23.56 & 43.05 & 30.10 \\
\hline
Qwen2.5-1.5B & 41.76 & 41.55 & 51.03 & 51.03 \\
Qwen2.5-3B & 33.42 & 28.22 & 43.62 & 33.43 \\
Qwen2.5-7B& 32.11 & 25.63 & 43.12 & 31.24 \\
Qwen2.5-14B & 65.33 & 63.42 & 53.25 & 42.62 \\
Qwen2.5-32B & 68.13 & 63.53 & 53.87 & 43.39 \\
Qwen2.5-72B & 32.25 & 26.34 & 43.57 & 34.20 \\
\hline
Gemma-2-2B & 30.82 & 23.56 & 43.06 & 30.11 \\
Gemma-2-9B & 54.37 & 54.37 & 50.99 & 50.65 \\
Gemma-2-27B & 69.39 & 63.95 & 55.58 & 41.72 \\
\hline
Phi-3-mini & 32.28 & 26.05 & 43.23 & 32.96 \\
Phi-3-small & 35.28 & 30.68 & 44.35 & 41.48 \\
Phi-3-medium & 40.77 & 38.12 & 44.19 & 42.75 \\
\hline
Mistral-7B-v0.3 & 38.13 & 36.23 & 46.37 & 45.53 \\
Mixtral-8x7B-v0.1 & 68.29 & 43.32 & 56.96 & 36.97 \\
Ministral-8B-2410 & 30.86 & 23.63 & 43.06 & 30.13 \\
\hline
\end{tabular}
\caption{The accuracy and F1 scores of different models in determining whether sub-questions should be retrieved or directly answered.}
\label{tab:ex2-part2}
\vspace{-0.3cm}
\end{table}

While evaluating LLMs' capability to solve questions using their parametric knowledge, we observed frequent  ``I don't know'' responses or no answer. This highlights the challenges LLMs face in solving complex multi-hop questions using only internal knowledge or limited single-step reasoning capabilities. To address these, LLM often incorporate sub-question decomposition and retrieval strategies. However, the effectiveness heavily depends on the model's ability to decide when to use them. We analyze this across three critical aspects.


\subsection{Direct Answer vs. Decompositions vs. Retrieval}
When encountering multi-hop questions, models must choose between direct answering, sub-question generation, or retrieval. This decision significantly impacts system efficiency and accuracy. Specifically, simple factual questions are often answered directly, while multi-hop or rare fact queries benefit from decomposition or retrieval. 


As shown in Table \ref{tab:ex2-part1}, Phi-3-small-8k performs best on \textsc{MINTQA-pop} (Accuracy: 75.79\%, F1: 55.56\%), while GPT-4o leads on \textsc{MINTQA-ti} (Accuracy: 65.46\%, F1: 38.18\%). However, model size doesn't always predict performance; Qwen2.5-32B underperforms its 14B variant. Lower-performing models, like Gemma-2-2B, favor direct answering (92.59\% on \textsc{MINTQA-ti}), likely due to their limited ability to assess question complexity.
\begin{table}[!t]
\tiny
\centering
\begin{tabular}{l|cc|cc}
\hline
\multirow{2}{*}{Model} & \multicolumn{2}{c|}{\textsc{MINTQA-pop}} & \multicolumn{2}{c}{\textsc{MINTQA-ti}} \\
\cline{2-5}
& Acc & F1 & Acc & F1 \\
\hline
GPT-3.5 & 37.74 & 30.78 & 43.74 & 38.20 \\
GPT-4o-mini & 59.31 & 58.96 & 56.25 & 56.02 \\
GPT-4o & 71.30 & 71.22 & 59.30 & 57.91 \\
\hline
LLaMA-3.1-8B & 34.81 & 25.82 & 40.83 & 28.99 \\
LLaMA-3.1-70B & 58.19 & 57.67 & 52.43 & 52.43 \\
LLaMA-3.2-1b & 23.14 & 20.09 & 33.86 & 23.47 \\
LLaMA-3.2-3B & 28.62 & 22.63 & 37.52 & 26.98 \\
\hline
Qwen2.5-1.5B & 34.81 & 25.82 & 40.83 & 28.99 \\
Qwen2.5-3B & 77.49 & 72.19 & 59.52 & 49.39 \\
Qwen2.5-7B & 62.16 & 62.13 & 53.25 & 52.85 \\
Qwen2.5-14B & 79.05 & 78.53 & 60.02 & 57.83 \\
Qwen2.5-32B & 95.94 & 95.52 & 62.53 & 58.74 \\
Qwen2.5-72B & 83.68 & 83.06 & 62.03 & 59.54 \\
\hline
Gemma-2-2B & 34.81 & 25.82 & 40.83 & 28.99 \\
Gemma-2-9B & 40.05 & 34.31 & 43.79 & 38.25 \\
Gemma-2-27B & 65.83 & 65.82 & 55.32 & 54.93 \\
\hline
Phi-3-mini & 34.87 & 25.92 & 41.11 & 30.07 \\
Phi-3-small & 47.19 & 44.44 & 46.71 & 44.78 \\
Phi-3-medium & 35.36 & 26.77 & 42.65 & 35.16 \\
\hline
Mistral-7B-v0.3 & 34.91 & 25.99 & 40.87 & 29.31 \\
Mixtral-8x7B-v0.1 & 35.96 & 28.18 & 41.73 & 32.52 \\
Ministral-8B-2410 & 34.81 & 25.82 & 40.88 & 29.14 \\
\hline
\end{tabular}
\caption{
The accuracy and F1 scores of the model in determining whether the main question has been answered based on the given sub-question-answer pair.
}
\label{tab:ex2-part3}
\vspace{-0.2cm}
\end{table}

\subsection{Direct Answer vs. Retrieval for Sub-questions}
\label{dir_2}
When handling sub-questions, models must decide between direct answering and retrieval based on the required knowledge. Popular facts might be answered directly, while tail knowledge or recent information often requires retrieval. 

Our experiments results in Table \ref{tab:ex2-part2}  reveal a general correlation between model size and decision quality, with some exceptions. LLaMA-3.1-70B outperforms other LLaMA variants, achieving 54.60\% and 48.46\% accuracy on \textsc{MINTQA-pop} and \textsc{MINTQA-ti}, respectively. However, GPT-4o underperforms GPT-3.5, likely due to overconfidence in its parametric knowledge, as it selects direct answering on 93.48\% of \textsc{MINTQA-pop} and 69.20\% of \textsc{MINTQA-ti} questions. Additionally, models perform better on \textsc{MINTQA-ti}, indicating new knowledge provides a clearer signal for retrieval compared to knowledge of varying popularity, where the decision boundary is less distinct.


\subsection{Decomposition vs. Synthesis}
\label{dir_3}

For multi-hop questions (hop count $\geq 2$), we evaluate models' ability to decide whether to decompose further or synthesize the final answer from intermediate results. As shown in Table \ref{tab:ex2-part3}, performance generally correlates with model size. Qwen2.5-32B achieves 95\% accuracy on \textsc{MINTQA-pop} but drops to 62.53\% on \textsc{MINTQA-ti}, reflecting new knowledge poses challenges for synthesizing. Some models like Mistral-7B, show extreme biases, predicting the main answer always within sub-answers for 99.90\% cases of \textsc{MINTQA-pop}.

\begin{figure}
    \centering
    \includegraphics[width=0.8\linewidth]{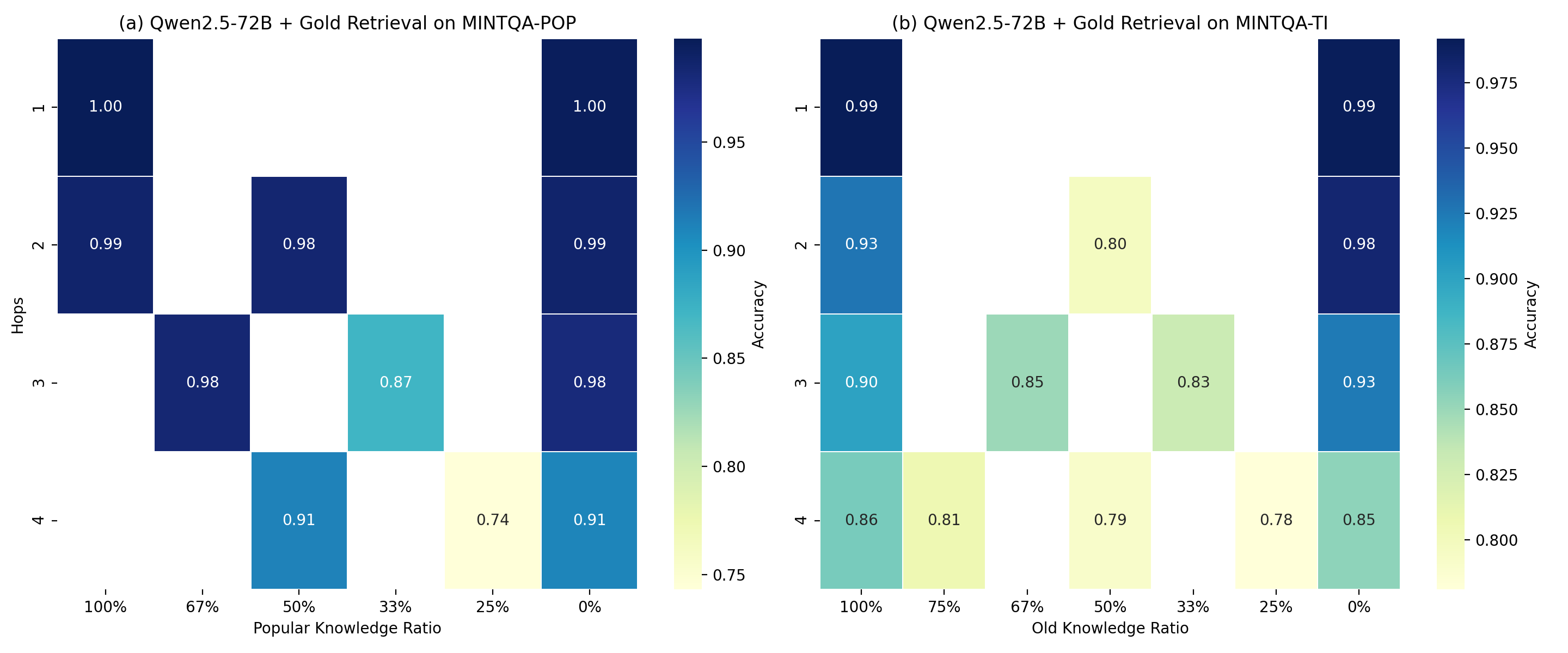}
    \caption{The performance of Qwen2.5-72B with gold retrieval across two datasets. The X-axis represents the proportion of popular knowledge required in the question, and the Y-axis indicates question hops.}
    \label{fig:direct_QA_gold_error}
    \vspace{-0.4cm}
\end{figure}

\begin{figure*}
    \centering
    \includegraphics[width=0.7\linewidth]{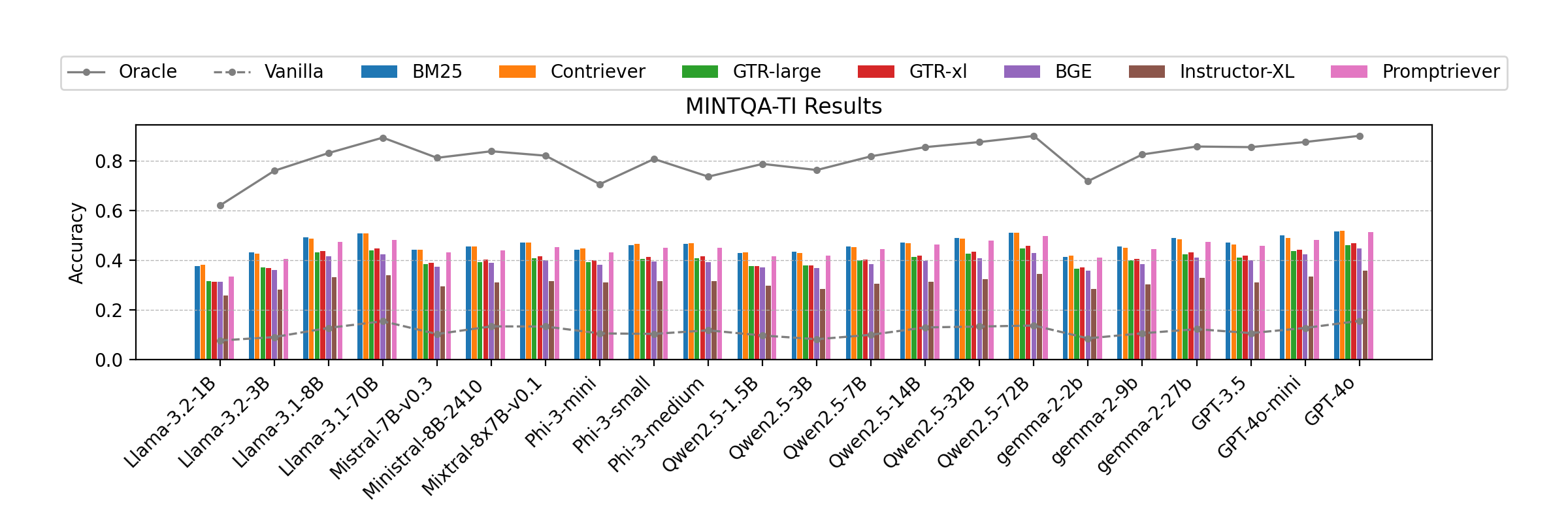}
    \caption{Performance comparison of LLMs on \textsc{MINTQA-ti} using different retrieval methods: "Oracle" uses gold-standard retrieval passages, while "Vanilla" involves models answering without retrieval content.}
    \label{fig:RAG_direct_results_TI}
    \vspace{-0.2cm}
\end{figure*}

\section{Effectiveness of Direct Retrieval for Multi-hop QA}
\label{sec7}
Having analyzed LLMs' performance using only their parametric knowledge on MINTQA in Section 5 and their decision-making capabilities in strategy selection in section 6, we now explore whether incorporating these strategic decisions can improve their performance. In this section, we evaluate the effectiveness of applying direct retrieval to LLMs to handle our complex multi-hop questions.



\subsection{Retrieval Model Selection and Configuration}
\textbf{Retrieval Models} We evaluate seven retrieval approaches across three categories:
1) Sparse retriever: \textbf{BM25} \cite{Robertson2009ThePR}. 2) Vector retrievers pre-trained on large unlabeled corpora: \textbf{Contriever} \cite{Izacard2021UnsupervisedDI}, \textbf{GTR-LARGE/XL} \cite{Ni2021LargeDE} and \textbf{BGE} \cite{Xiao2023LMCocktailRT}. 3) Instruction-tuned text embedding retrievers:
\textbf{Instructor-XL} \cite{INSTRUCTOR} and \textbf{Promptriever} \cite{Weller2024PromptrieverIR}..

\noindent \textbf{Configuration} 
We follow the approach of \citet{YuZNZL0HWWX23} to construct the retrieval corpus by linearizing the knowledge graph $\mathcal{G}$ into text. $\mathcal{G}$ consists of $\mathcal{G}_{pop}$ (Section \ref{sec:MINTQA-pop_construction}) and $\mathcal{G}_{ti}$ (Section \ref{sec:MINTQA-ti_construction}). See Appendix \ref{app:KG_linearization} for details.

\subsection{Performance Analysis}

Following prior work \cite{Mallen2022WhenNT,Maekawa2024RetrievalHO}, we use direct retrieval method to enhance LLM performance and evaluate them. For each retrieval, we select the top-5 passages that are relevant to the question and input them as context. Our analysis reveals both potential and limitations of this approach.


Figure \ref{fig:RAG_direct_results_TI} demonstrates that retrieval significantly enhances performance, especially on \textsc{MINTQA-ti}, with an average 30\% accuracy gain over the Vanilla setting (no retrieval). Similar trends are observed on \textsc{MINTQA-pop} (refer to Figure \ref{fig:RAG_direct_results}). Notably, in the Oracle setting, where gold-standard passages are used, even small models like Llama-3.2-1B achieve a 25\% accuracy improvement compared to the average performance of all retrievers we used, emphasizing the potential for better retrievers. Appendix \ref{app:AnalysisofDirectRetrieval} provides more analysis of the retriever.

We analyze the impact of knowledge popularity and newness on QA performance. Models with different retrievers show inconsistent patterns when varying proportions of new and popular knowledge. To isolate retrieval quality, we pair models with gold retrieval. Figures \ref{fig:direct_QA_gold_error}(a) and (b) show that with Qwen2.5-72B with gold retrieval, performance initially declines and then improves as the proportion of popular or old knowledge decreases. This likely occurs because the model effectively determine whether using parametric knowledge and retrieval for fully familiar (100\% popular/old) or unfamiliar (100\% unpopular/new) questions but struggles with mixed knowledge, leading to errors. Further analyses are in Appendix \ref{app:AnalysisofDirectRetrieval}.

\section{Effectiveness of Sub-Question Generation for Multi-hop QA}
\label{sec8}
In this section, we investigate whether generating and answering sub-questions or providing sub-questions for answering improves the accuracy on our benchmark. Results can be seen in Table \ref{tab:self-generate}.

\paragraph{Self-Generated Sub-Questions:}  
On \textsc{MINTQA-pop}, self-generated sub-questions improve performance slightly (e.g., LLaMA-3.1-8B: 34.83\% to 37.28\%), but they degrade accuracy on \textsc{MINTQA-ti} (12.83\% to 9.28\%). This contrast reflects the reliance on models' knowledge bases: for known but unpopular facts in \textsc{MINTQA-pop}, decomposition organizes existing knowledge, while on \textsc{MINTQA-ti}, knowledge gaps might lead to flawed decomposition and additional errors.

\paragraph{Providing Sub-Questions:}  
Gold sub-questions significantly boost performance on \textsc{MINTQA-pop} (e.g., LLaMA-3.1-8B sees a 33.41\% increase) by clarifying reasoning paths and allowing models to focus on synthesis. On \textsc{MINTQA-ti}, improvements are modest, with the best accuracy (23.75\%) still from LLaMA-3.1-8B. This differences can be expected. While decomposition can help models better utilize their existing knowledge, it cannot compensate for the fundamental lack of information when handling questions about new facts.

\section{Enhancing Multi-hop QA through Integrating Decomposition and Retrieval}
\label{sec9}
In practice, solving multi-hop questions often requires combining these question decomposition and retrieval effectively. This section explores two integration strategies and provides oracle analysis.

\subsection{Decomposition-then-Retrieval Approach}

\begin{table}[!t]
\scriptsize
\centering
\begin{tabular}{l|cc|cc}
\hline
\multirow{2}{*}{Model}      & \multicolumn{2}{c}{\textsc{MINTQA-pop}} & \multicolumn{2}{c}{\textsc{MINTQA-ti}} \\
\cline{2-5}

           & (1) & (2)                     & (1) & (2)                     \\
\hline
LLaMA-3.2-1B & 19.62 & 41.89 &6.88 &  14.23 \\
LLaMA-3.2-3B & 27.02 &  59.93 &7.42 & 18.26 \\
LLaMA-3.1-8B & 37.28 &  70.69 &9.82 & 23.75 \\
LLaMA-3.1-70B & 54.08 & 72.97 &16.73 &  23.15 \\
\hline
Mistral-7B-v0.3 & 31.41 &  58.80 &9.52 & 16.00 \\
Ministral-8B-2410 & 35.76 &  59.61 &10.54 & 17.44 \\
Mixtral-8x7B-v0.1 & 45.66 & 68.90 &12.08 &  20.93 \\
\hline
Phi-3-mini & 21.41 &  37.61 & 5.84 &11.46 \\
Phi-3-small & 26.98 & 38.53 & 8.09 & 13.52 \\
Phi-3-medium & 38.91 &  64.29 & 10.18 &19.45 \\
\hline
Qwen2.5-1.5B & 25.74 & 59.29 &9.07 &  18.33 \\
Qwen2.5-3B & 23.86 &  55.90 &6.99 & 15.87 \\
Qwen2.5-7B & 26.32 & 55.83 & 8.93 & 17.85 \\
Qwen2.5-14B & 41.34 & 65.67 &12.96 &  20.01 \\
Qwen2.5-32B & 39.16 &  63.44 &12.51 & 19.79 \\
Qwen2.5-72B & 44.99 &  54.62 &14.10 & 20.51 \\
\hline
Gemma-2-2B & 25.67 & 49.97 &8.04 &  14.32 \\
Gemma-2-9B & 38.18 &59.65 &  11.17 & 18.09 \\
Gemma-2-27B & 44.54 &66.29 &  12.36 & 18.90 \\
\hline
\end{tabular}
\caption{The accuracy of LLMs evaluated under query decomposition settings: (1) the model generates and answers sub-questions itself, and (2) the model answers given gold sub-questions.}
\label{tab:self-generate}
\vspace{-0.3cm}
\end{table}
Building on prior work \cite{Li2023SelfAdaptiveRO, Shi2024GeneratethenGroundIR}, we implement an iterative decomposition-then-retrieval approach for multi-hop QA. At each step, the LLM decides whether to decompose the question further or synthesize a final answer using previous sub-question results. If yes,  5 relevant documents are retrieved for the new sub-question, with up to five iterations allowed. Each step incorporates the full history of sub-questions and answers as context; if no, it synthesizes previous results for all subquestions into a final answer. We evaluate with three top-performing retrieval approaches: BM25, Contriever, and Promptriever\footnote{GPT models were excluded due to high cost and limited performance advantages over open-source LLMs (70B+)}.

Figure \ref{fig:rag_decom_QA} shows performance varies across datasets. On \textsc{MINTQA-pop}, larger models (>14B) benefit from decomposition and retrieval compared to direct retrieval, while smaller models (<8B) perform worse due to decomposition errors. On \textsc{MINTQA-ti}, decomposition-then-retrieval underperform direct retrieval for most models, suggesting new knowledge poses greater challenges for question decomposition.

\subsection{Oracle Analysis with Gold Component}
We evaluate system limitations using gold-standard sub-questions and their retrieved documents. In this oracle setting (i.e. Gold subqeustion + Gold retrieval in Figure \ref{fig:rag_decom_QA}), models achieve over 90\% accuracy across both datasets, including the challenging \textsc{MINTQA-ti}. Figure \ref{fig:rag_decom_QA} further shows notable gains across all models and retrievers when using gold sub-questions, especially for smaller LLMs,  highlighting their difficulties in generating accurate sub-questions independently.

However, even with perfect decomposition and relevant documents, accuracy remains below 100\%. This reveals two challenges: extracting relevant information from documents containing multiple facts and synthesizing information across sub-questions, suggesting areas for future improvement beyond retrieval and decomposition.

\begin{table}[!t]
\centering
\tiny
\begin{tabular}{l|ccc}
\hline
\multirow{2}{*}{Model} & \multicolumn{3}{c}{BM25} \\
\cline{2-4}
& Acc (\%) & Avg. Sub & Avg. Ret  \\
\hline
\multicolumn{4}{c}{\textsc{MINTQA-pop}} \\
\hline
\multicolumn{4}{l}{\textbf{Qwen Models}} \\
Qwen2.5\mbox{-}1.5B & 25.86 & 1.13 & 28.31  \\
Qwen2.5\mbox{-}14B & 53.77 & 3.44 & 35.56 4 \\
Qwen2.5\mbox{-}32B & 50.33 & 2.79 & 42.27  \\
Qwen2.5\mbox{-}3B & 31.16 & 1.78 & 86.53  \\
Qwen2.5\mbox{-}72B & 58.63 & 3.01 & 44.76  \\
Qwen2.5\mbox{-}7B & 32.98 & 2.18 & 52.29  \\
\hline
\multicolumn{4}{l}{\textbf{Llama Models}} \\
LLaMA\mbox{-}3.1\mbox{-}70B & 64.80 & 3.41 & 99.61  \\
LLaMA\mbox{-}3.1\mbox{-}8B & 50.01 & 3.88 & 97.57  \\
LLaMA\mbox{-}3.2\mbox{-}1B & 20.53 & 1.79 & 76.01  \\
LLaMA\mbox{-}3.2\mbox{-}3B& 37.23 & 3.48 & 93.65  \\
\hline
\multicolumn{4}{l}{\textbf{Mistral Models}} \\
Mistral\mbox{-}7B\mbox{-}v0.3 & 29.47 & 3.13 & 58.72  \\
Mixtral\mbox{-}8x7B\mbox{-}v0.1 & 48.28 & 3.61 & 35.81  \\
Ministral\mbox{-}8B\mbox{-}\mbox{-}2410 & 35.91 & 2.95 &0.55\\
\hline
\multicolumn{4}{l}{\textbf{Phi Models}} \\
Phi\mbox{-}3\mbox{-}medium & 40.16 & 2.98 & 36.31  \\
Phi\mbox{-}3\mbox{-}mini& 26.81 & 4.75 & 46.98 \\
Phi\mbox{-}3\mbox{-}small & 37.09 & 2.92 & 26.66 \\
\hline
\multicolumn{4}{l}{\textbf{Gemma Models}} \\
Gemma\mbox{-}2\mbox{-}27B& 64.64 & 4.05 & 98.89 \\
Gemma\mbox{-}2\mbox{-}2B & 34.20 & 4.96 & 19.88 \\
Gemma\mbox{-}2\mbox{-}9B & 39.99 & 3.93 & 8.17  \\
\hline
\end{tabular}
\caption{The results for ``Generate then Adaptively Retrieve'' are as follows: \textbf{Acc} represents the accuracy of the model in answering questions, \textbf{Avg. Sub} indicates the average number of sub-questions generated by the model, and \textbf{Avg. Ret} refers to the average number of sub-questions deemed necessary for retrieval by the model.}
\label{tab:dynamic-generate-partial}
\vspace{-0.3cm}
\end{table}

\subsection{Decomposition-Dynamic Retrieval Approach}
The iterative decomposition-then-retrieve strategy in Section 9.1 faces two key challenges: high computational overhead from repeated retrievals \cite{Zhuang2024EfficientRAGER} and performance degradation from unnecessary retrievals \cite{Mallen2022WhenNT, Maekawa2024RetrievalHO}. To address this, we explore whether LLMs can dynamically determine retrieval necessity. Following \cite{Ni2024WhenDL}, we implement a confidence-guided retrieval mechanism, triggering retrieval only when models express low confidence in answering sub-questions (details in App. \ref{appdix_exp}). Table \ref{tab:dynamic-generate-partial} shows some results, with complete results in App. \ref{compelet_results}.

Our analysis reveals two key findings. First, reducing retrievals while maintaining performance proves challenging, with only the largest models (LLaMA-3.1-70B and Gemma-2-27B) maintaining accuracy despite high retrieval rates (>98\%). Other models show significant performance drops, reflecting our datasets' emphasis on rare and new information. Second, Models exhibit varying retrieval dependencies. Mistral and Phi models show high self-confidence (55\% retrieval rate), LLaMA variants consistently trigger retrieval (>90\%), while Gemma models exhibit size-dependent behavior, with retrieval rates ranging from <10\% (2-9B) to >98\% (2-27B) on \textsc{MINTQA-pop}.

\section{Conclusion}
In this work, we introduce \textbf{MINTQA}, a multi-hop QA benchmark reasoning across popular/unpopular and old/new knowledge. MINTQA spans reasoning chains from one to four hops, enabling systematic assessment of LLMs' complex reasoning abilities. We also propose a comprehensive evaluation framework to assess key aspects of multi-hop QA, including question handling strategy selection, the effectiveness of question decomposition and retrieval, which allows detailed analysis of models' decision-making and reasoning capabilities. Evaluations on state-of-the-art LLMs reveal that even the best LLM with retrieval still struggle on our benchmark. Our analysis highlights the limitations of LLMs 
in multi-hop QA, providing insights to advance LLMs' reasoning capabilities in complex multi-hop QA.



\section*{Limitation}

This work has several key limitations. First, our definition of long-tail and new facts relies solely on Wikidata distribution patterns, which may not very accurately reflect knowledge representation in LLMs' diverse pre-training corpora. Second, our simplified approach to constructing the retrieval corpus by concatenating entity-related facts into sequential sentences—differs from the complexity of real-world documents and might potentially overestimate the performance of retrieval-augmented methods. Third, budget constraints limited our evaluation of powerful closed-source models like GPT-4, though preliminary results suggest our benchmark remains challenging even for these advanced systems.
Regarding methodology, while our prompting strategy proved effective on the sampled data, we did not explore advanced techniques such as iterative prompt optimization or chain-of-thought prompting \cite{Wei2022ChainOT}. However, we hypothesize that such optimizations would yield limited improvements, as the core challenge lies in models' knowledge gaps.

\bibliography{acl_latex_arxiv}
\clearpage
\appendix
\section{Data Analysis}
\label{data_analy_app}
\begin{figure}
    \centering
    \includegraphics[width=1.1\linewidth]{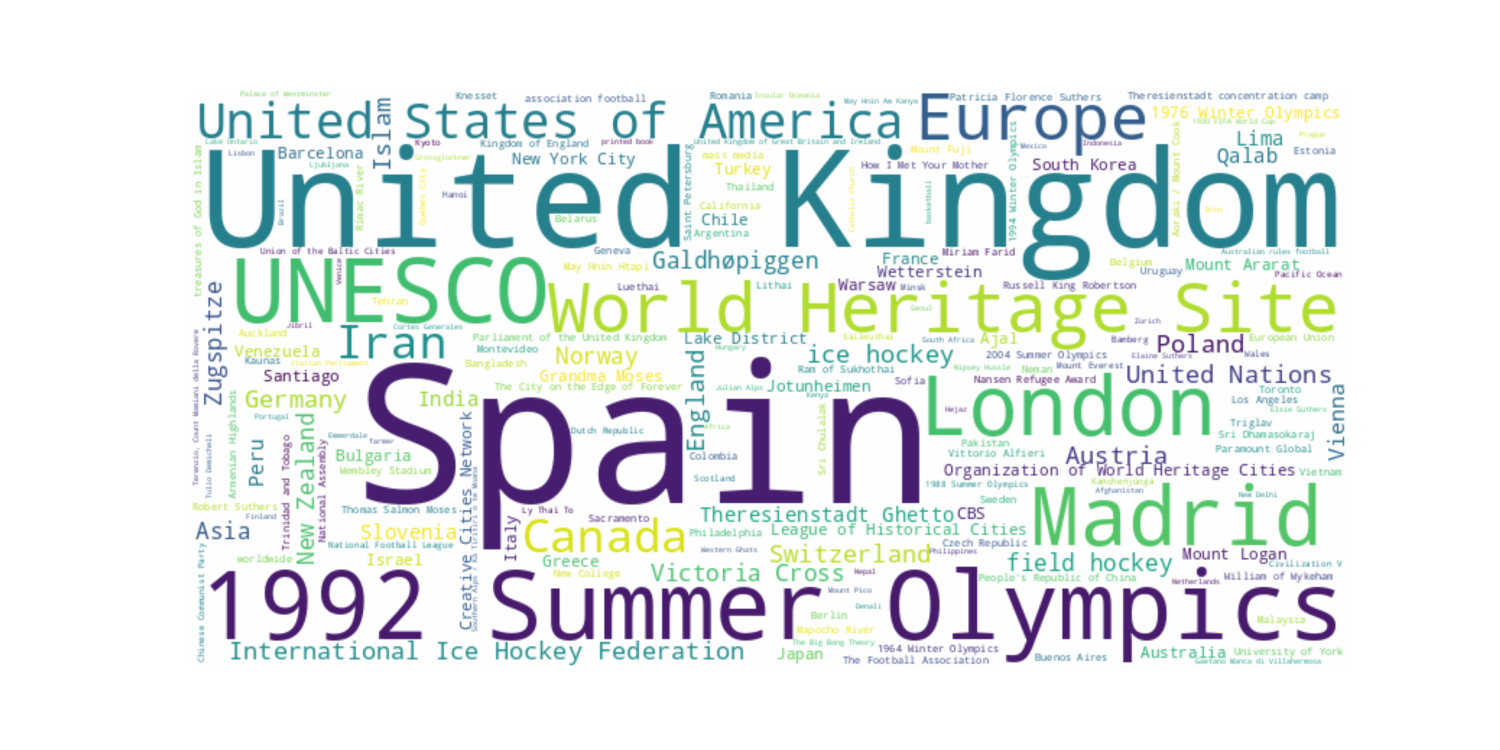}
    \caption{The word cloud of the entities.}
    \label{fig:ent_cloud}
\end{figure}
\begin{figure}
    \centering
    \includegraphics[width=1.1\linewidth]{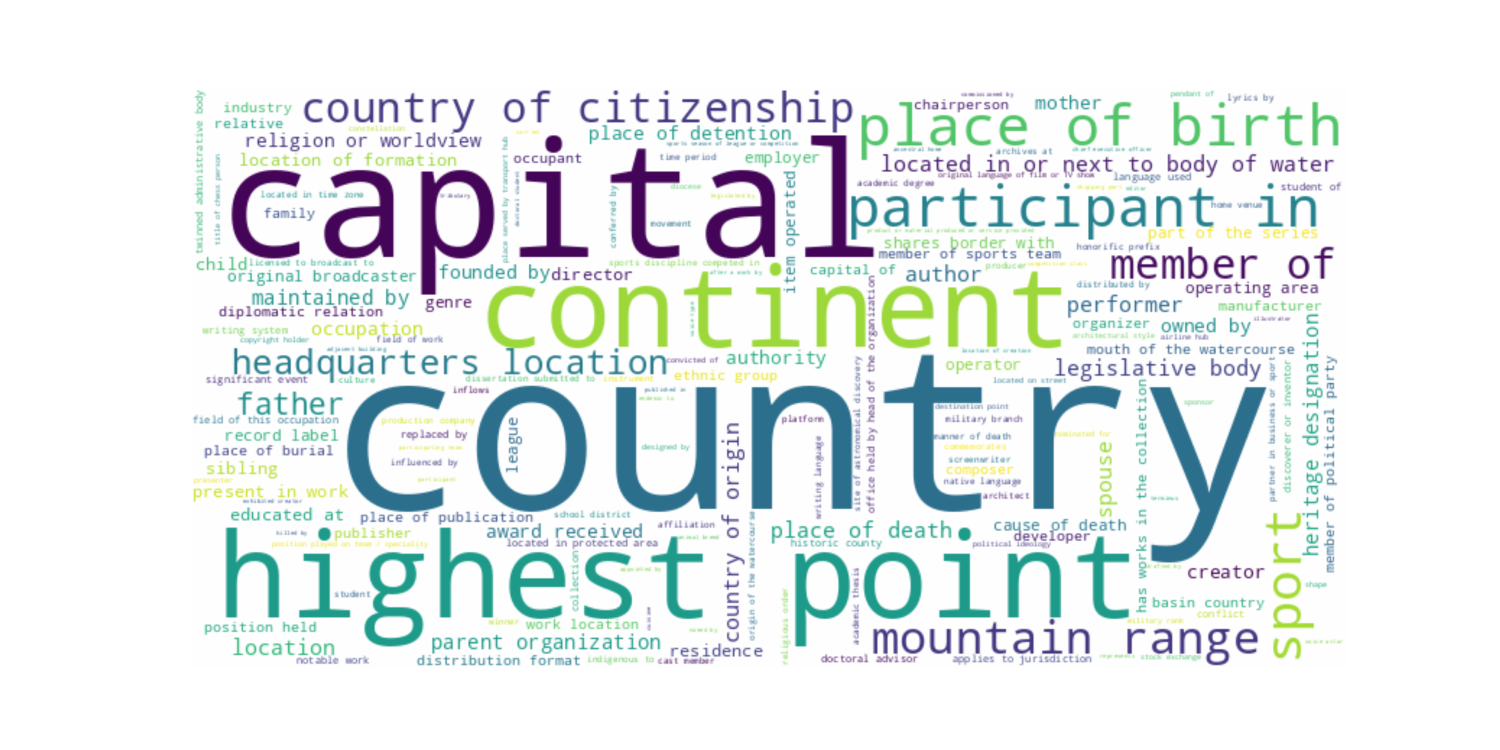}
    \caption{The word cloud of the relations.}
    \label{fig:re_cloud}
\end{figure}
\subsection{Word Cloud Distribution}
We conducted a word cloud analysis on the triplets used in our dataset. We can observe from Figure \ref{fig:ent_cloud} and \ref{fig:re_cloud} that the most frequent entities in our constructed dataset are related to geographical information, with ``United Kingdom'' appearing the most frequently. Following that, there are entities related to events, such as ``1992 Summer Olympics''.

Regarding relationships, the word ``Country'' appears the most, followed closely by ``Capital''. These relationships are also related to geographical information. However, we can observe other noticeably frequent relationships, such as those related to ``Place of Birth'' and ``Participant In'', which are associated with individuals and events.

\begin{figure}[t]
    \centering
    \includegraphics[width=1\linewidth]{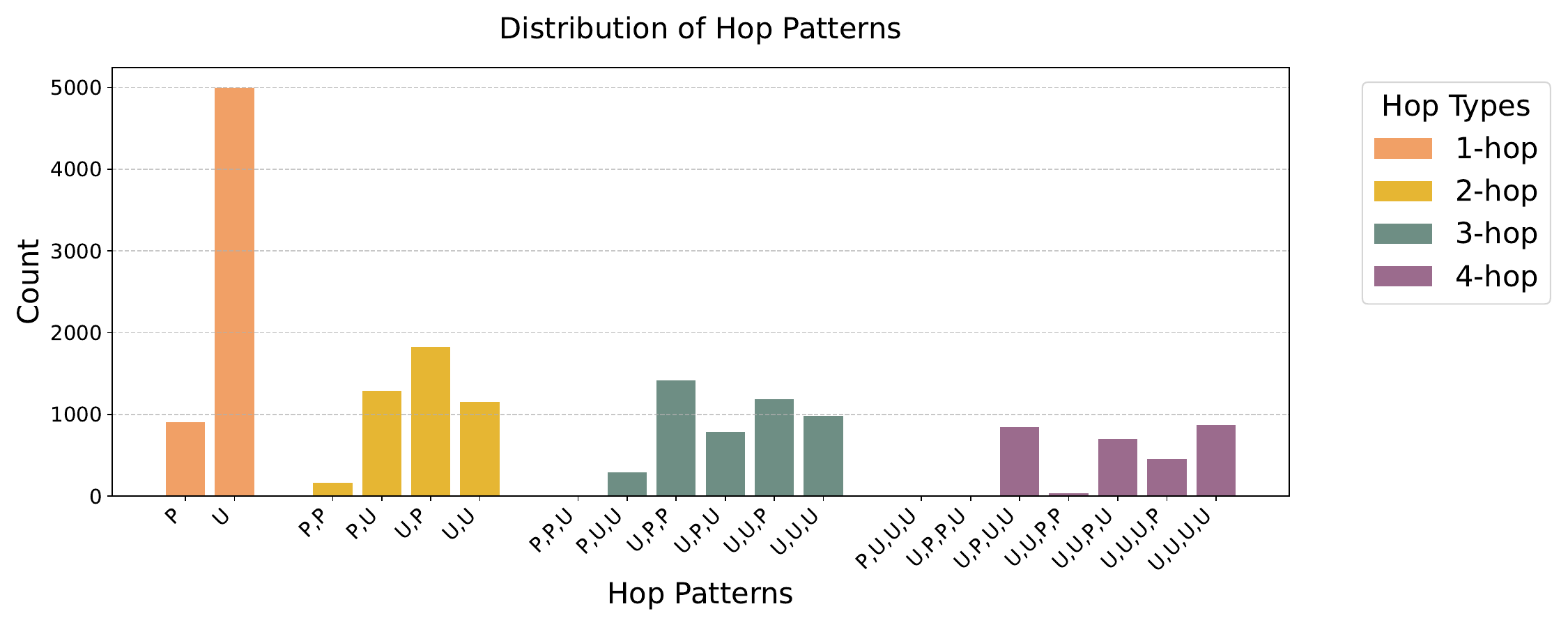}
    \caption{Popularity Related Data Distribution}
    \label{fig:data_statis_pop}
\end{figure}
\begin{figure}[t]
    \centering
    \includegraphics[width=1\linewidth]{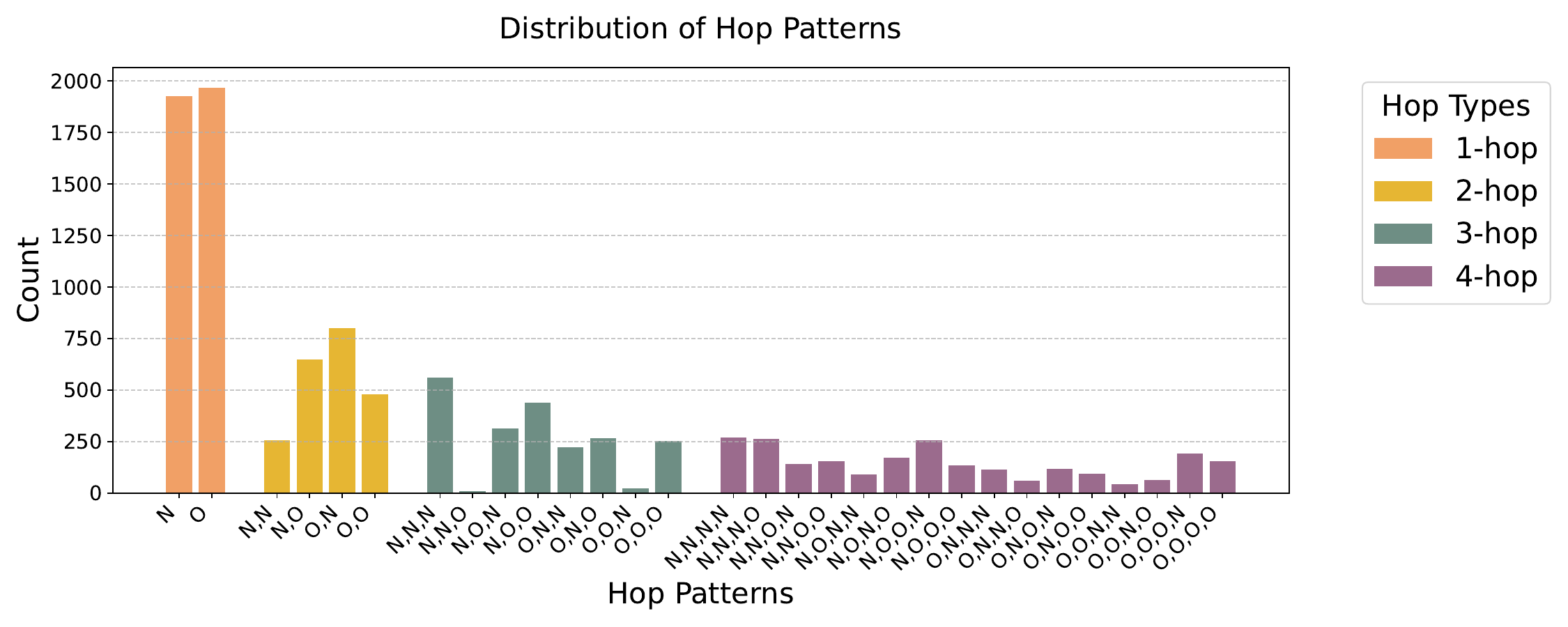}
    \caption{Time Related Data Distribution}
    \label{fig:data_statis_ti}
\end{figure}
\subsection{Data Type Distribution}
Figures \ref{fig:data_statis_pop} and \ref{fig:data_statis_ti} show the distributions of popular versus unpopular facts and old versus new facts within each hop category. Due to our focus on model performance with unpopular and new facts, we sampled more of these fact types. Certain fact combinations such as ``P,P,P'', three-hop chains composed entirely of popular facts, do not occur in our dataset, so they are not shown in the figures. 


\section{Human Sampling Check}
\begin{table*}[t]
\scriptsize
    \centering
    \begin{tabular}{c|ccccc|c}
    \hline
    Dataset& \# Questions & Multi-hop & Time & Popularity & Sub-questions & Original of Generated Questions \\
    \hline
    PopQA \cite{Mallen2022WhenNT} & 14,268 & $\times$ & $\times$ & $\checkmark$ &$\times$ & Templates \\
    WiTQA \cite{Maekawa2024RetrievalHO} & 14,837 & $\times$ & $\times$ & $\checkmark$ & $\times$ &Machine \\
    Head-to-tail \cite{Sun2023HeadtoTailHK}& 18,171 & $\times$ & $\times$ & $\checkmark$ &$\times$ & Template \\
    RetrievalQA \cite{Zhang2024RetrievalQAAA} & 1,271 & $\times$ & $\checkmark$ & $\checkmark$ & $\times$ &Mixed \\
    \hline 
    FreshQA \cite{vu-etal-2024-freshllms} & 600 & $\checkmark$ & $\checkmark$ & $\times$ &$\times$ & Human \\
    MultihopQA-RAG \cite{Tang2024MultiHopRAGBR} & 2,556 & $\checkmark$ & $\times$ & $\times$ &$\times$ & Machine \\
    HotpotQA \cite{Yang2018HotpotQAAD} & 112,779 & $\checkmark$ & $\times$ & $\times$ &$\times$ & Human \\
    MuSiQue \cite{Trivedi2021MM} & 24,814 & $\checkmark$ & $\times$ & $\times$ &$\times$ & Human \\
    2WikiMultiHopQA \cite{Ho2020ConstructingAM} & 192,606 & $\checkmark$ & $\times$ & $\times$ & $\times$ &Templates \\
    Mintaka \cite{Sen2022MintakaAC} & 20,000 & $\checkmark$ & $\times$ & $\times$ & $\times$ &Human \\
    FanoutQA \cite{Zhu2024FanOutQAAM}& 1,034 & $\checkmark$ & $\times$ & $\times$ & $\checkmark$  &Human \\
    \hline 
    Ours & 28,366 & $\checkmark$ & $\checkmark$ & $\checkmark$ &$\checkmark$  & Machine \\
    \hline
    \end{tabular}
\caption{Comparison between our dataset and other datasets.}
\label{data_compare}
\end{table*}

Despite our data filtering strategy's success in improving data quality compared to the initial dataset, some failure cases still exist. Our analysis revealed that certain samples in the MINTQAQA dataset share a common limitation: they contain reasoning questions that lack sufficient context for definitive answers. For instance, questions like ``Which event did the Iberian Revolutionary Liberation Directory participate in?'' demonstrate this issue. 

Given MINTQA's role as an evaluation benchmark, we took measures to understand the effects of such cases. We implemented a rigorous human validation process. Two specialists who are English native speakers were hired to conduct a systematic evaluation of our randomly selected 500 instances from MINTQA. Our primary objective was to verify whether each question could be answered definitively with the provided information. The annotators were tasked with assessing whether each question could be answered unambiguously based on the available context. 
The results were highly encouraging: only 2\% of the evaluated samples exhibited contextual insufficiency, and no other significant issues were identified. These findings validate MINTQA's overall quality while also confirming the effectiveness of our sample filtering methodology. This low error rate demonstrates that our quality control pipeline successfully maintains the dataset's integrity and reliability for evaluation purposes.

\section{Details of Benchmark}
\label{data_creation}
\subsection{Details of Benchmark Curation}
In Section \ref{sec3}, we present a comprehensive description of our benchmark construction methodology. Our approach includes carefully designed prompts for both question generation and validation processes. The complete specifications of these prompts are illustrated in figures 9 through 15.

\subsection{License}
Our benchmark data are released under the MIT License, which is detailed in https:// opensource.org/licenses/MIT. 

\section{Comparison with Existing Benchmarks}
In this section, we provide a comprehensive comparison with question answering benchmarks closely related to our own in Table \ref{data_compare}.  

Compared to previous benchmarks, ours encompasses both old/new knowledge and unpopular/popular knowledge, presenting new challenges for retrieval-augmented large language model systems. Furthermore, unlike RetrievalQA \cite{Zhang2024RetrievalQAAA}, which covers old/new or unpopular/popular knowledge but relies on integrating existing QA datasets, our benchmark generates questions using language models, enabling scalable data construction. RetrievalQA, on the other hand, is constrained by the limited availability of existing datasets and focuses solely on short-form open-domain question answering.  

Additionally, while multi-hop datasets exist, only FreshQA \cite{Vu2023FreshLLMsRL} involves new knowledge in questions. However, FreshQA's data is manually created, limited to just 600 samples, and lacks scalability. Our dataset, by contrast, provides sub-questions that assist in evaluating or training models on intermediate reasoning steps in multi-hop processes, enabling a more comprehensive assessment of LLMs' capabilities on similar tasks.  

This more integrated benchmark can help the research community gain deeper insights into the weaknesses of large models in question answering, improve training methods, and address the limitations of current benchmark practices.

%

\section{Qualitative Analysis}
\label{case_study}
Table~\ref{first_case} to Table \ref{last-case} present representative examples of multi-hop questions and their corresponding sub-questions generated by GPT-4o for both \textsc{MINTQA-pop} and \textsc{MINTQA-ti} datasets. We have selected three representative instances for each hop level, ranging from single-hop to four-hop questions. As demonstrated in the table, GPT-4o effectively converted the triplets into well-structured, coherent questions. The high quality of these generated questions makes them suitable for evaluating retrieval-augmented LLMs' capabilities in handling multi-hop questions that involve rare and new knowledge.

\begin{table*}[!tbp]
    \centering
    \tiny
    \begin{tabular}{l|cccccc|ccccccc}
        \hline
        \multirow{2}{*}{Model} & \multicolumn{6}{c|}{POP} & \multicolumn{7}{c}{TI} \\
        \cline{2-14}
        & 1 & 0.67 & 0.5 & 0.33 & 0.25 & 0 & 1 & 0.75 & 0.67 & 0.5 & 0.33 & 0.25 & 0 \\        \hline
        \multicolumn{14}{l}{\textbf{GPT}} \\
        \hline
        GPT-3.5 & 83.94 & 2.74 & 69.59 & 18.46 & 9.40 & 49.34 & 9.68 & 4.34 & 8.79 & 8.88 & 2.02 & 2.30 & 17.97 \\
        GPT-4o & 89.11 & 3.23 & 87.83 & 29.95 & 14.65 & 63.42 & 14.73 & 7.85 & 15.80 & 14.03 & 3.85 & 6.91 & 22.89 \\
        GPT-4o-Mini & 84.32 & 2.88 & 78.55 & 26.24 & 12.30 & 47.96 & 12.31 & 6.20 & 12.09 & 11.32 & 3.49 & 4.93 & 18.90 \\
        \hline
        \multicolumn{14}{l}{\textbf{Llama}} \\
        \hline
        Llama-3.1 & 75.87 & 2.18 & 54.91 & 19.88 & 7.70 & 38.29 & 11.29 & 5.37 & 8.65 & 9.59 & 5.50 & 5.43 & 21.73 \\
        Llama-3.1-70B & 90.42 & 2.32 & 68.99 & 22.84 & 12.15 & 55.23 & 13.47 & 7.44 & 11.95 & 15.80 & 4.22 & 7.57 & 23.06 \\
        Llama-3.2-1B & 64.51 & 0.49 & 29.30 & 4.24 & 5.45 & 23.59 & 6.35 & 2.69 & 5.36 & 5.33 & 2.57 & 2.96 & 14.29 \\
        Llama-3.2-3B & 75.87 & 1.41 & 58.91 & 14.22 & 7.05 & 29.17 & 7.40 & 5.37 & 6.18 & 6.84 & 3.12 & 4.44 & 15.91 \\
        \hline
        \multicolumn{14}{l}{\textbf{Qwen}} \\
        \hline
        Qwen-1.5B & 62.44 & 3.79 & 49.13 & 14.84 & 7.00 & 22.68 & 8.49 & 5.17 & 8.10 & 7.32 & 2.39 & 4.44 & 16.41 \\
        Qwen-3B & 62.82 & 3.65 & 51.57 & 12.06 & 6.40 & 23.40 & 7.89 & 4.96 & 5.22 & 5.68 & 3.30 & 4.28 & 13.62 \\
        Qwen-7B & 73.62 & 5.34 & 61.49 & 12.81 & 6.35 & 26.87 & 9.40 & 4.55 & 7.83 & 7.81 & 3.12 & 4.93 & 16.15 \\
        Qwen-14B & 78.59 & 5.20 & 73.66 & 37.68 & 11.75 & 35.18 & 13.29 & 6.82 & 12.36 & 10.39 & 4.77 & 5.43 & 18.97 \\
        Qwen-32B & 81.13 & 6.04 & 76.87 & 30.17 & 12.25 & 35.69 & 12.56 & 7.23 & 13.46 & 11.45 & 3.49 & 7.89 & 19.47 \\
        Qwen-72B & 80.56 & 5.41 & 78.74 & 31.85 & 10.60 & 40.55 & 12.63 & 8.47 & 14.29 & 11.90 & 3.67 & 5.26 & 20.63 \\
        \hline
        \multicolumn{14}{l}{\textbf{Gemma}} \\
        \hline
        Gemma-2-2B & 57.18 & 1.05 & 36.83 & 10.73 & 4.40 & 24.39 & 7.30 & 4.13 & 5.08 & 5.73 & 1.47 & 2.80 & 16.18 \\
        Gemma-2-9B & 80.28 & 1.62 & 65.33 & 26.63 & 11.75 & 33.99 & 10.73 & 4.55 & 6.04 & 7.19 & 2.39 & 4.28 & 18.14 \\
        Gemma-2-27B & 80.56 & 2.74 & 70.07 & 27.83 & 9.85 & 37.74 & 12.17 & 6.40 & 9.20 & 9.50 & 3.85 & 7.40 & 18.87 \\
        \hline
        \multicolumn{14}{l}{\textbf{Phi}} \\
        \hline
        Phi-3-mini-4K & 79.53 & 2.67 & 64.41 & 23.06 & 10.70 & 33.17 & 9.61 & 5.58 & 11.13 & 9.19 & 2.39 & 7.57 & 15.45 \\
        Phi-3-small-8K & 74.74 & 2.18 & 60.06 & 21.33 & 9.60 & 34.34 & 9.93 & 5.17 & 7.42 & 9.41 & 2.02 & 5.26 & 15.85 \\
        Phi-3-medium-4K & 84.79 & 2.60 & 70.35 & 25.57 & 11.40 & 45.97 & 10.87 & 6.82 & 12.50 & 11.10 & 2.20 & 4.28 & 17.34 \\
        \hline
        \multicolumn{14}{l}{\textbf{Mistral}} \\
        \hline
        Mistral-7B & 81.31 & 1.48 & 48.78 & 12.81 & 8.00 & 36.20 & 9.12 & 3.93 & 7.14 & 7.55 & 2.39 & 4.11 & 18.21 \\
        Mixtral-8x7B & 84.69 & 2.74 & 66.03 & 24.16 & 10.95 & 47.30 & 13.43 & 5.58 & 10.85 & 10.92 & 4.22 & 5.10 & 20.40 \\
        Mistral-8B & 76.06 & 4.50 & 56.34 & 28.75 & 11.75 & 35.57 & 10.84 & 5.58 & 11.26 & 9.28 & 5.32 & 8.72 & 23.29 \\
        \hline
    \end{tabular}
        \caption{The model's accuracy in the zero-shot setting is analyzed within \textsc{MINTQA-pop} and \textsc{MINTQA-ti}, categorized based on the proportion of popular facts and old facts. A value of 0 indicates that the questions are entirely composed of unpopular facts or new facts, with other numbers increasing proportionally.}

    \label{tab:app_zero}
\end{table*}

\section{Additional Experimental Details}
\label{appdix_exp}
\subsection{Implementation Details}
In our experiments, we utilized the following state-of-the-art LLMs, with detailed version specifications: GPT-3.5 (gpt-3.5-turbo-1106), GPT4o-min (gpt-4o-mini-2024-07-18), GPT4o (gpt-4o-2024-08-06), LLaMA-3.1-8B (LLaMA-3.1-8B-instruct), LLaMA-3.1-70B (LLaMA-3.1-70B-instruct), LLaMA-3.2-1B (LLaMA-3.2-1B-instruct), LLaMA-3.2-3B (LLaMA-3.2-3B-Instruct), Qwen-2.5-1.5B (Qwen-2.5-1.5B-Instruct), Qwen-2.5-3B (Qwen-2.5-3B-Instruct), Qwen-2.5-7B (Qwen-2.5-7B-Instruct), Qwen-2.5-14B (Qwen-2.5-14B-Instruct), Qwen-2.5-32B (Qwen-2.5-32B-Instruct), Qwen-2.5-72B (Qwen-2.5-72B-Instruct), Gemma-2-2b (Gemma-2-2b-it), Gemma-2-9b (Gemma-2-9b-it), Gemma-2-27b (Gemma-2-27b-it), Phi-3-mini (Phi-3-mini-4k), Phi-3-small (Phi-3-small-8k), Phi-3-medium (Phi-3-medium-4k), Mistral-7b (mistral-7b-instruct-v0.3), Mixtral-8X7b (Mixtral-8X7B-instruct-v0.1), and Ministral-8B (Ministral-8B-instruct-2410). All experiments were conducted using 4 A100 (80GB) GPUs.
From Table \ref{tab:question-prompt} to \ref{confidence_task_prompt}, we provide the prompts used to instruct these models in completing their respective tasks.

\subsection{Retrievers and KG Linearization Details}
\label{app:KG_linearization}
We evaluate seven retrieval approaches across three categories:
1) Sparse retriever: \textbf{BM25} \cite{Robertson2009ThePR}. 2) Vector retrievers pre-trained on large unlabeled corpora: \textbf{Contriever} \cite{Izacard2021UnsupervisedDI}: Fine-tuned on MS-MARCO, \textbf{GTR-LARGE/XL} \cite{Ni2021LargeDE} and \textbf{BGE} \cite{Xiao2023LMCocktailRT}: Further fine-tuned on NQ \cite{Kwiatkowski2019NaturalQA} and HotpotQA \cite{Yang2018HotpotQAAD}. 3) Instruction-tuned text embedding retrievers:
\textbf{Instructor-XL} \cite{INSTRUCTOR}: Multi-task trained on 330 tasks for instruction robustness. \textbf{Promptriever} \cite{Weller2024PromptrieverIR}: Uses LLaMA backbone, trained on curated instance-level instruction sets from MS-MARCO, demonstrating superior retrieval performance compared to Instructor-XL.

We linearise the knowledge graph (KG) $\mathcal{G}$ as a source of text retrieval in the corpus, with reference to the work in \citet{YuZNZL0HWWX23}. Specifically, for each entity in $\mathcal{G}$, we extract a 1-hop subgraph centered on the entity and convert it into linearized text, treating it as a passage. Since $\mathcal{G}$ includes both old and new versions of the Wikidata dump, knowledge conflicts may arise due to updates. Conflicting triples are separated into different passages. Each passage is split into chunks of 512 tokens, a size shown to be effective for practical applications \cite{wang2024searching}.

\section{Additional Experiments and Result Analysis}

\subsection{Zero-shot: Performance Across Retrieval Categories}
\label{app_zero_shot}

In Table \ref{tab:app_zero}, we present the performance of large language models in a zero-shot evaluation setting across different proportions of unpopular/popular and old/new facts. As observed, the accuracy is highest when questions are composed solely of popular or old facts. For example, LLaMA-3.1-70B achieves an accuracy of 90.42\% on \textsc{MINTQA-pop} and 13.47\% on \textsc{MINTQA-ti}.

However, as the proportion of unpopular or new facts increases, the accuracy of the models shows a declining trend. Interestingly, when this proportion reaches 1, the accuracy tends to rise compared to lower ratios. This is likely because the proportion of 1 often includes many 1-hop questions, which are comparatively easier for the models to resolve.

\begin{table}[t]
    \small
\begin{tabular}{l|cc|cc}
\hline
\multirow{2}{*}{Model} & \multicolumn{2}{c|}{Class A} & \multicolumn{2}{c}{Class B} \\
\cline{2-5}
& Acc. & F1 & Acc. & F1 \\
\hline
\multicolumn{5}{l}{\textbf{GPT}} \\
\hline
GPT-3.5 & 89.4 & 52.1 & 31.5 & 46.2 \\
GPT-4o & 99.7 & 49.4 & 9.3 & 16.9 \\
GPT-4o-Mini & 97.3 & 53.5 & 25.7 & 40.5 \\
\hline
\multicolumn{5}{l}{\textbf{Llama}} \\
\hline
Llama-3.1 & 100.0 & 47.1 & 0.1 & 0.2 \\
Llama-3.1-70B & 94.9 & 56.3 & 36.6 & 52.8 \\
Llama-3.2-1B & 99.3 & 47.1 & 1.2 & 2.3 \\
Llama-3.2-3B & 100.0 & 47.1 & 0.0 & 0.0 \\
\hline
\multicolumn{5}{l}{\textbf{Qwen}} \\
\hline
Qwen-1.5B & 57.9 & 38.0 & 34.6 & 45.1 \\
Qwen-3B & 97.9 & 47.5 & 4.7 & 8.9 \\
Qwen-7B & 100.0 & 47.6 & 1.9 & 3.7 \\
Qwen-14B & 68.9 & 55.1 & 63.7 & 71.8 \\
Qwen-32B & 52.9 & 50.6 & 74.9 & 76.5 \\
Qwen-72B & 98.2 & 47.2 & 2.8 & 5.5 \\
\hline
\multicolumn{5}{l}{\textbf{Gemma}} \\
\hline
Gemma-2-2B & 100.0 & 47.1 & 0.0 & 0.0 \\
Gemma-2-9B & 88.7 & 54.5 & 39.1 & 54.2 \\
Gemma-2-27B & 49.6 & 50.0 & 78.2 & 78.0 \\
\hline
\multicolumn{5}{l}{\textbf{Phi}} \\
\hline
Phi-3-Mini-4K & 99.5 & 47.5 & 2.4 & 4.6 \\
Phi-3-Small-8K & 99.0 & 48.5 & 6.9 & 12.8 \\
Phi-3-Medium-4K & 99.7 & 50.9 & 14.5 & 25.3 \\
\hline
\multicolumn{5}{l}{\textbf{Mistral}} \\
\hline
Mistral-7B & 89.8 & 47.2 & 15.1 & 25.2 \\
Mixtral-8x7B & 3.1 & 5.7 & 97.3 & 80.9 \\
Mistral-8B & 100.0 & 47.1 & 0.1 & 0.1 \\
\hline
\end{tabular}
\caption{The per-label accuracy and F1 scores for the tasks of sub-question judgment, retrieval, or direct answer generation. }
\label{app_ex2_part2}
\end{table}
\begin{table}[t]
\small
\begin{tabular}{l|cc|cc}
\hline
\multirow{2}{*}{Model} & \multicolumn{2}{c|}{Class A} & \multicolumn{2}{c}{Class B} \\
\cline{2-5}
& Acc. & F1 & Acc. & F1 \\
\hline
\multicolumn{5}{l}{\textbf{GPT}} \\
\hline
GPT-3.5 & 99.7 & 52.7 & 4.6 & 8.8 \\
GPT-4o & 95.2 & 69.8 & 58.5 & 72.7 \\
GPT-4o-Mini & 98.5 & 62.8 & 38.4 & 55.1 \\
\hline
\multicolumn{5}{l}{\textbf{Llama}} \\
\hline
Llama-3.1 & 100.0 & 51.6 & 0.0 & 0.0 \\
Llama-3.1-70B & 99.6 & 62.4 & 36.1 & 53.0 \\
Llama-3.2-1B & 100.0 & 51.6 & 0.0 & 0.0 \\
Llama-3.2-3B & 100.0 & 51.6 & 0.0 & 0.0 \\
\hline
\multicolumn{5}{l}{\textbf{Qwen}} \\
\hline
Qwen-1.5B & 100.0 & 51.6 & 0.0 & 0.0 \\
Qwen-3B & 48.6 & 60.1 & 92.9 & 84.3 \\
Qwen-7B & 93.4 & 63.2 & 45.5 & 61.0 \\
Qwen-14B & 91.1 & 75.2 & 72.6 & 81.9 \\
Qwen-32B & 93.7 & 94.1 & 97.1 & 96.9 \\
Qwen-72B & 92.7 & 79.8 & 78.9 & 86.3 \\
\hline
\multicolumn{5}{l}{\textbf{Gemma}} \\
\hline
Gemma-2-2B & 100.0 & 51.6 & 0.0 & 0.0 \\
Gemma-2-9B & 100.0 & 53.7 & 8.1 & 14.9 \\
Gemma-2-27B & 97.1 & 66.4 & 49.1 & 65.2 \\
\hline
\multicolumn{5}{l}{\textbf{Phi}} \\
\hline
Phi-3-Mini-4K & 100.0 & 51.7 & 0.1 & 0.2 \\
Phi-3-Small-8K & 99.7 & 56.8 & 19.1 & 32.1 \\
Phi-3-Medium-4K & 100.0 & 51.8 & 0.9 & 1.7 \\
\hline
\multicolumn{5}{l}{\textbf{Mistral}} \\
\hline
Mistral-7B & 100.0 & 51.7 & 0.2 & 0.3 \\
Mixtral-8x7B & 98.9 & 51.8 & 2.3 & 4.6 \\
Mistral-8B & 100.0 & 51.6 & 0.0 & 0.0 \\
\hline
\end{tabular}
\caption{The per-label accuracy and F1 scores for the task where the model is required to determine whether the answer to the main question has been found, given the sub-questions and their answers.}
\label{app_ex2_part3}

\end{table}
\subsection{Accuracy and F1 Across Categories}
Table \ref{app_ex2_part2} and \ref{app_ex2_part3} reports the accuracy and F1 scores for each category under the evaluation setup described in Section \ref{dir_2} and \ref{dir_3}. From the table, we can observe that most models demonstrate high accuracy, often exceeding 90\% or even reaching 100\% in identifying sub-questions that can directly generate answers. However, the F1 scores are significantly lower. This discrepancy indicates that models tend to predict that all examples are solvable, revealing an overconfidence in their ability to answer our constructed benchmarks.

The table also highlights similar phenomena across models, particularly for LLaMA-3.1-8B, LLaMA-3.2-1B, LLaMA-3.2-3B, Qwen2.5-1.5B, Gemma-2-2B, and Mistral-8B. These models consistently predict that the main question can be derived from existing sub-question answers. On the other hand, models in the same series, such as Qwen2.5 variants, exhibit more balanced accuracy and F1 scores across categories. This reflects significant inconsistencies among large models in determining whether sub-question answers suffice to answer the main question.

Such findings indicate the challenges of relying on large models for complex reasoning tasks and highlight the need for more robust evaluation metrics and methodologies.

\subsection{Sub-question Generation Analysis}
From Figures \ref{fig:matrix}, we illustrate the relationship between the number of sub-questions generated by models and the corresponding gold sub-question counts. This analysis considers scenarios where models are required to independently generate and answer sub-questions.

We observe substantial differences among models of similar sizes. For instance, the Qwen2.5-7B model tends to generate fewer sub-questions, with most counts falling in the range of 1 or 2. In contrast, the Mistral-7B model produces sub-questions with a more uniform distribution, primarily ranging from 2 to 5. Despite these differences, smaller models, such as Qwen2.5-1.5B and LLaMA-3.2-1B, exhibit similar trends. Both predominantly generate only 1 sub-question, reflecting the limited capability of these smaller LLMs to generate sub-questions as part of their answering process.
Examining the distributions of larger models on the \textsc{MINTQA-pop} and \textsc{MINTQA-ti} datasets reveals that, despite differences in the datasets, large models exhibit similar distributions in terms of actual step counts and the number of sub-questions generated by the models.

\subsection{More Analysis of Direct Retrieval}
\label{app:AnalysisofDirectRetrieval}

\begin{figure}
    \centering
    \includegraphics[width=1\linewidth]{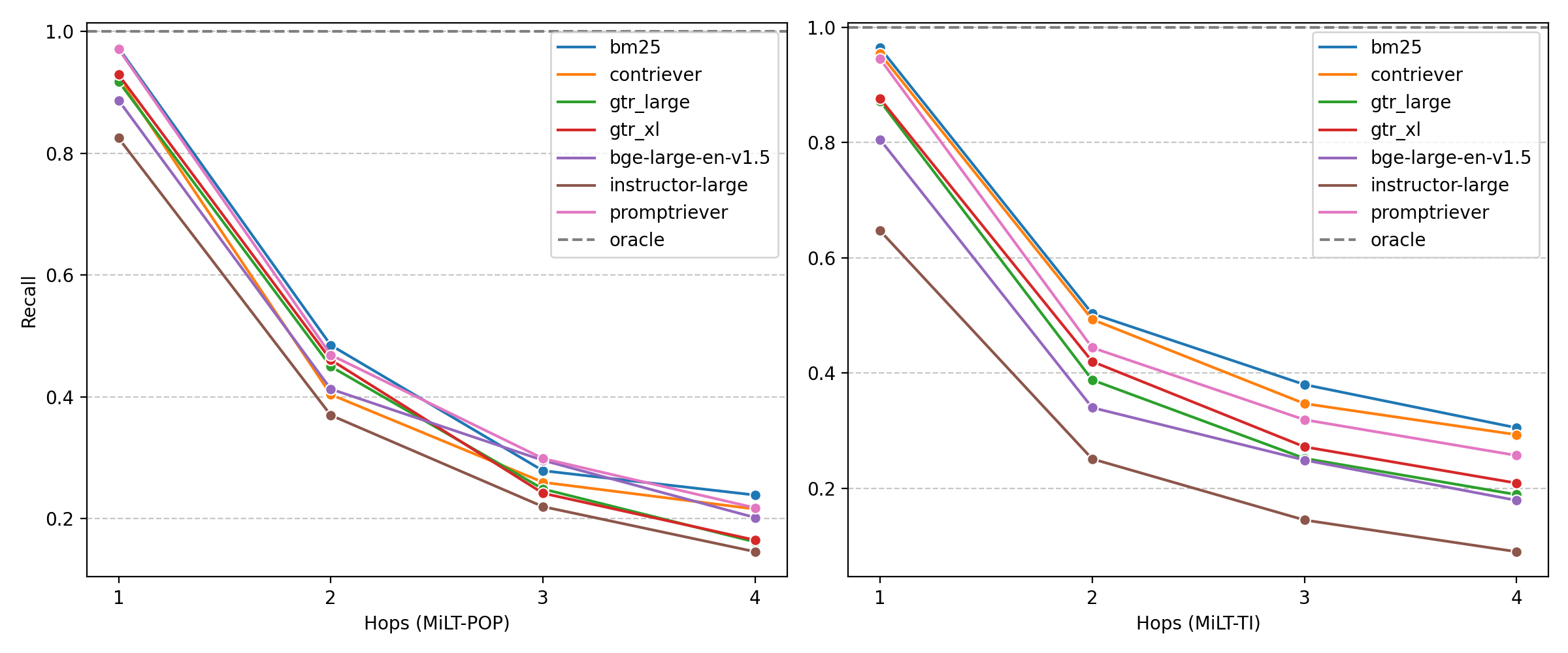}
    \caption{Recall performance of retrieval methods across two datasets for varying question hops.}
    \label{fig:retrieval_by_hops}
\end{figure}

Direct retrieval strategy have limitations when handling multi-hop questions. Figure \ref{fig:retrieval_by_hops} reveals significant limitations in current direct retrieval approaches when handling multi-hop questions. First, the retrieval effectiveness decreases markedly with increased hop count. We observe a consistent decline in recall rates across all retrieval methods as question complexity increases, indicating fundamental limitations in the direct retrieval approach. Second, among the retrieval methods evaluated, BM25 demonstrated the best performance. This can be explained by the highly structured nature of our KG-linearized corpus. While dense retrieval methods excel at capturing semantic similarities in natural text, BM25's lexical matching approach is well-suited for  knowledge graph-derived text.

\begin{figure}
    \centering
    \includegraphics[width=1\linewidth]{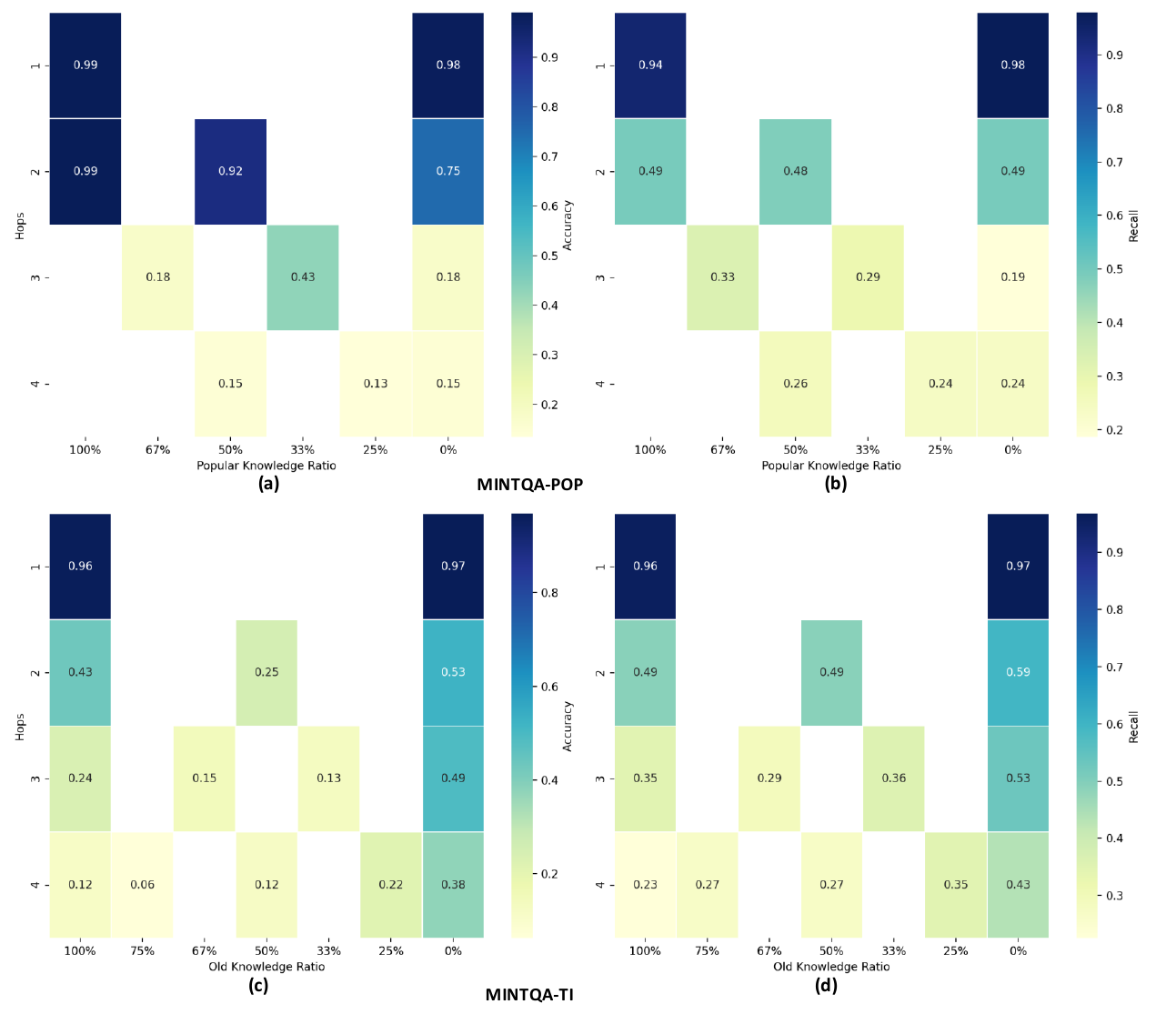}
    \caption{Heatmaps (a) and (c) show Qwen2.5-72B with BM25 performance on two datasets, while heatmaps (b) and (d) shows BM25 recall. The X-axis represents the proportion of popular knowledge required in the question, and the Y-axis indicates question hops.}
    \label{fig:direct_QA_pop_error}
\end{figure}

We demonstrate the influence of knowledge newness and popularity on direct retrieval scenarios, using Qwen2.5-72B paired with BM25 as a representative example. As shown in Figure \ref{fig:direct_QA_pop_error} (a) and (c), QA performance declines with an increasing proportion of unpopular or new knowledge in questions. However, performance improves when the proportion of new knowledge reaches 100\% (i.e., no old knowledge), as higher new knowledge presence boosts recall rates (Figure \ref{fig:direct_QA_pop_error} (d)), ultimately enhancing QA accuracy on \textsc{MINTQA-ti}. This highlights the retriever's effectiveness in handling new knowledge.
\subsection{Complete Results for Decomposition-Dynamic Retrieval}
\label{compelet_results}
Table \ref{tab:dynamic-generate-full} presents the complete results on \textsc{MINTQA-pop} and \textsc{MINTQA-ti} using large models to output confidence scores for sub-questions and determine whether retrieval is needed based on the confidence values. We conducted experiments with three retrievers: BM25, Contrieve, and PromptRetrieve.

\begin{figure}
    \centering
    \includegraphics[width=1\linewidth]{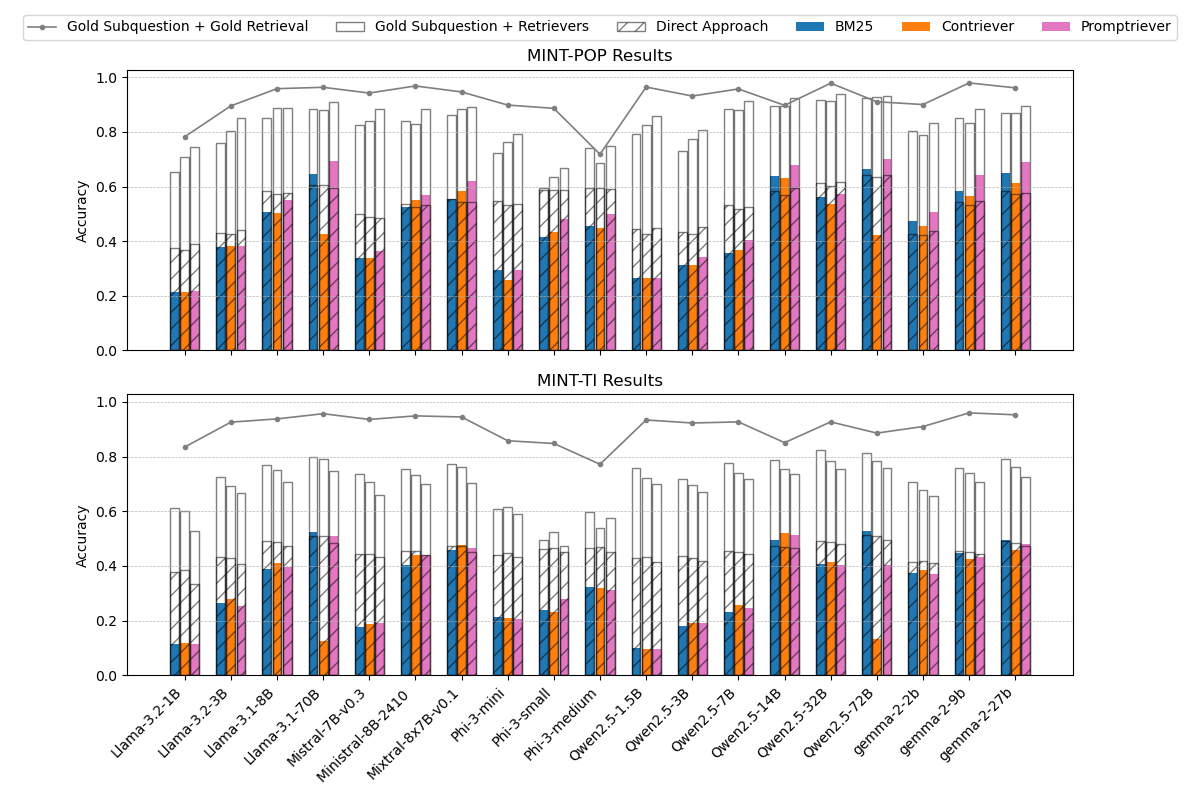}
    \caption{Performance of all models with the three retriever using decomposition-retrieval approach on two datasets.}
    \label{fig:rag_decom_QA}
\end{figure}

\begin{figure*}
    \centering
    \includegraphics[width=0.85\linewidth]{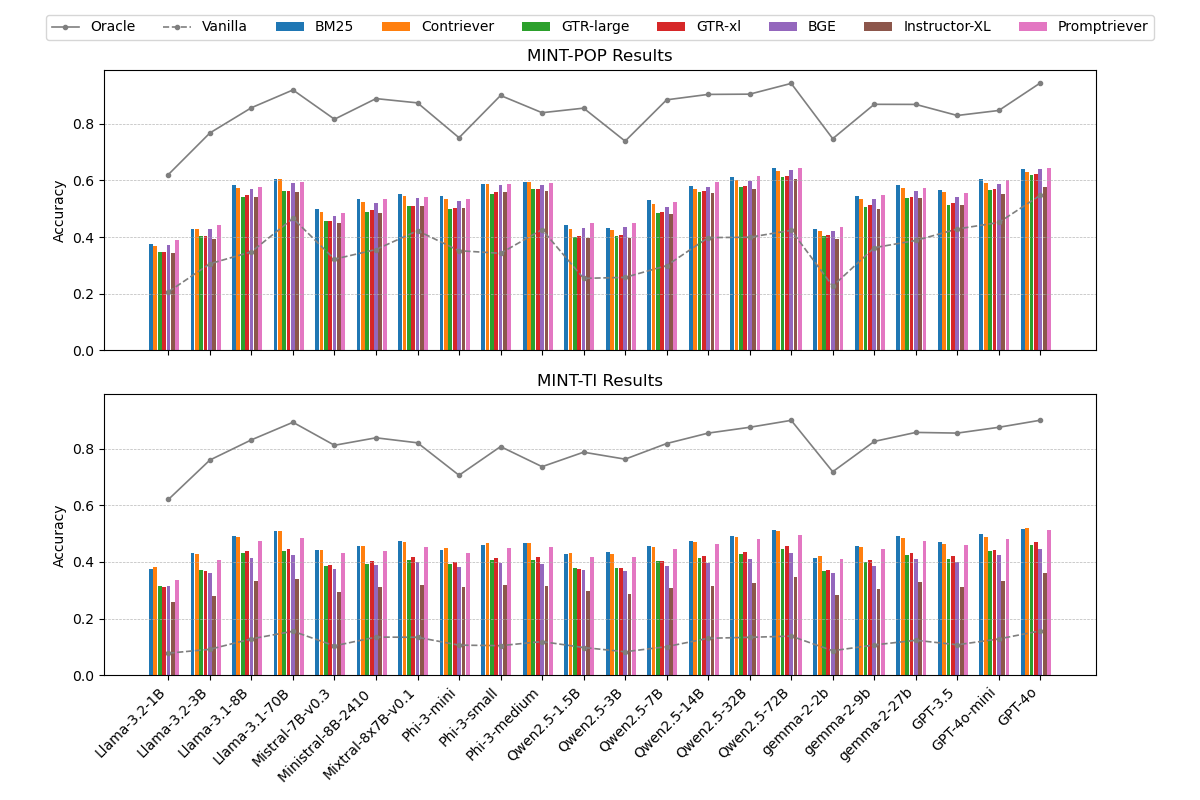}
    \caption{Performance comparison of LLMs on \textsc{MINTQA-pop} and \textsc{MINTQA-ti} using different retrieval methods. "Oracle" uses gold-standard retrieval passages, while "Vanilla" involves models answering without retrieval content.}
    \label{fig:RAG_direct_results}
\end{figure*}

\begin{table*}[ht]
\centering
\tiny
\begin{tabular}{l|ccc|ccc|ccc}
\hline
\multirow{2}{*}{Model} & \multicolumn{3}{c|}{BM25} & \multicolumn{3}{c|}{Contrieve} & \multicolumn{3}{c}{PromptRetrieval} \\
\cline{2-10}
& Acc (\%) & Avg. Sub & Avg. Ret  & Acc (\%) & Avg. Sub  & Avg. Ret  & Acc (\%) & Avg. Sub  & Avg. Ret  \\
\hline
\multicolumn{10}{c}{\textsc{MINTQA-pop}} \\
\hline
\multicolumn{10}{l}{\textbf{Qwen Models}} \\
Qwen2.5\mbox{-}1.5B & 25.86 & 1.13 & 28.31 & 26.02 & 1.15 & 27.33 & 26.13 & 1.15 & 27.44 \\
Qwen2.5\mbox{-}14B & 53.77 & 3.44 & 35.56 & 53.35 & 3.45 & 35.51 & 55.53 & 3.41 & 35.34 \\
Qwen2.5\mbox{-}32B & 50.33 & 2.79 & 42.27 & 48.56 & 2.77 & 43.04 & 50.84 & 2.81 & 42.07 \\
Qwen2.5\mbox{-}3B & 31.16 & 1.78 & 86.53 & 29.56 & 1.70 & 87.18 & 31.55 & 1.69 & 87.04 \\
Qwen2.5\mbox{-}72B & 58.63 & 3.01 & 44.76 & 57.42 & 3.05 & 44.47 & 60.89 & 3.02 & 43.66 \\
Qwen2.5\mbox{-}7B & 32.98 & 2.18 & 52.29 & 34.89 & 2.11 & 51.84 & 36.58 & 2.09 & 52.85 \\
\hline
\multicolumn{10}{l}{\textbf{Llama Models}} \\
LLaMA\mbox{-}3.1\mbox{-}70B & 64.80 & 3.41 & 99.61 & 62.55 & 3.38 & 99.63 & 69.21 & 3.31 & 99.61 \\
LLaMA\mbox{-}3.1\mbox{-}8B & 50.01 & 3.88 & 97.57 & 50.19 & 4.05 & 97.67 & 54.38 & 4.01 & 97.52 \\
LLaMA\mbox{-}3.2\mbox{-}1B & 20.53 & 1.79 & 76.01 & 20.86 & 1.79 & 73.50 & 21.13 & 1.80 & 73.37 \\
LLaMA\mbox{-}3.2\mbox{-}3B& 37.23 & 3.48 & 93.65 & 37.70 & 3.49 & 93.88 & 38.04 & 3.50 & 94.03 \\
\hline
\multicolumn{10}{l}{\textbf{Mistral Models}} \\
Mistral\mbox{-}7B\mbox{-}v0.3 & 29.47 & 3.13 & 58.72 & 28.80 & 3.24 & 54.67 & 30.73 & 3.20 & 57.56 \\
Mixtral\mbox{-}8x7B\mbox{-}v0.1 & 48.28 & 3.61 & 35.81 & 48.05 & 3.60 & 36.16 & 49.00 & 3.59 & 34.18 \\
Ministral\mbox{-}8B\mbox{-}\mbox{-}2410 & 35.91 & 2.95 & 0.55 & 36.04 & 2.94 & 0.53 & 36.04 & 2.94 & 0.51 \\
\hline
\multicolumn{10}{l}{\textbf{Phi Models}} \\
Phi\mbox{-}3\mbox{-}medium & 40.16 & 2.98 & 36.31 & 39.65 & 2.94 & 36.35 & 41.90 & 2.96 & 36.12 \\
Phi\mbox{-}3\mbox{-}mini& 26.81 & 4.75 & 46.98 & 25.64 & 4.71 & 50.50 & 27.63 & 4.71 & 48.51 \\
Phi\mbox{-}3\mbox{-}small & 37.09 & 2.92 & 26.66 & 37.19 & 2.88 & 27.43 & 39.14 & 2.87 & 26.84 \\
\hline
\multicolumn{10}{l}{\textbf{Gemma Models}} \\
Gemma\mbox{-}2\mbox{-}27B& 64.64 & 4.05 & 98.89 & 60.82 & 4.09 & 98.82 & 68.64 & 4.32 & 99.07 \\
Gemma\mbox{-}2\mbox{-}2B & 34.20 & 4.96 & 19.88 & 34.62 & 4.96 & 20.11 & 35.62 & 4.96 & 19.98 \\
Gemma\mbox{-}2\mbox{-}9B & 39.99 & 3.93 & 8.17 & 40.20 & 3.92 & 8.57 & 40.56 & 3.92 & 8.30 \\
\hline
\multicolumn{10}{c}{\textsc{MINTQA-ti}} \\
\hline
\multicolumn{10}{l}{\textbf{Qwen Models}} \\
Qwen2.5\mbox{-}1.5B & 9.49 & 1.13 & 31.17 & 9.39 & 1.14 & 31.49 & 9.47 & 1.14 & 31.53 \\
Qwen2.5\mbox{-}14B & 39.40 & 3.65 & 54.94 & 41.53 & 3.63 & 55.21 & 40.91 & 3.62 & 54.82 \\
Qwen2.5\mbox{-}32B & 33.87 & 2.86 & 65.51 & 34.91 & 2.87 & 65.70 & 33.75 & 2.89 & 65.35 \\
Qwen2.5\mbox{-}3B & 18.18 & 1.83 & 90.41 & 17.55 & 1.78 & 90.75 & 17.76 & 1.75 & 90.76 \\
Qwen2.5\mbox{-}72B & 44.79 & 3.42 & 59.75 & 45.99 & 3.45 & 59.77 & 44.44 & 3.47 & 59.68 \\
Qwen2.5\mbox{-}7B & 20.36 & 2.35 & 67.42 & 21.83 & 2.21 & 69.75 & 21.34 & 2.21 & 70.45 \\
\hline
\multicolumn{10}{l}{\textbf{Llama Models}} \\
LLaMA\mbox{-}3.1\mbox{-}70B & 51.99 & 3.70 & 99.46 & 52.16 & 3.69 & 99.39 & 50.59 & 3.71 & 99.37 \\
LLaMA\mbox{-}3.1\mbox{-}8B & 38.24 & 3.99 & 97.73 & 40.40 & 4.10 & 97.78 & 39.02 & 4.11 & 97.72 \\
LLaMA\mbox{-}3.2\mbox{-}1B & 9.25 & 1.94 & 72.61 & 9.47 & 1.98 & 67.81 & 8.99 & 1.99 & 68.19 \\
LLaMA\mbox{-}3.2\mbox{-}3B & 25.89 & 3.73 & 93.03 & 27.39 & 3.75 & 93.19 & 24.82 & 3.74 & 93.05 \\
\hline
\multicolumn{10}{l}{\textbf{Mistral Models}} \\
Mistral\mbox{-}7B\mbox{-}v0.3 & 9.86 & 3.19 & 53.88 & 9.77 & 3.41 & 51.53 & 10.16 & 3.36 & 53.89 \\
Mixtral\mbox{-}8x7B\mbox{-}v0.1 & 24.91 & 3.97 & 33.44 & 26.19 & 3.95 & 33.41 & 25.72 & 3.96 & 32.48 \\
Ministral\mbox{-}8B\mbox{-}2410 & 10.69 & 2.99 & 3.27 & 10.75 & 2.99 & 3.21 & 10.71 & 2.98 & 3.23 \\
\hline
\multicolumn{10}{l}{\textbf{Phi Models}} \\
Phi\mbox{-}3\mbox{-}medium & 19.58 & 3.19 & 42.20 & 19.12 & 3.21 & 41.70 & 19.18 & 3.19 & 41.80 \\
Phi\mbox{-}3\mbox{-}mini & 17.38 & 4.68 & 52.05 & 16.94 & 4.65 & 56.79 & 17.21 & 4.65 & 53.90 \\
Phi\mbox{-}3\mbox{-}small & 15.38 & 3.15 & 39.14 & 15.71 & 3.11 & 39.23 & 16.20 & 3.11 & 38.83 \\
\hline
\multicolumn{10}{l}{\textbf{Gemma Models}} \\
Gemma\mbox{-}2\mbox{-}27B & 48.39 & 4.49 & 98.61 & 45.35 & 4.53 & 98.65 & 47.45 & 4.53 & 98.65 \\
Gemma\mbox{-}2\mbox{-}2B & 24.67 & 4.93 & 36.10 & 25.34 & 4.94 & 36.90 & 24.48 & 4.93 & 36.54 \\
Gemma\mbox{-}2\mbox{-}9B & 15.29 & 4.59 & 19.91 & 16.08 & 4.58 & 19.95 & 16.50 & 4.58 & 19.99 \\
\hline
\end{tabular}
\caption{The full results for ``Generate then Adaptively Retrieve'' are as follows: \textbf{Acc} represents the accuracy of the model in answering questions, \textbf{Avg. Sub} indicates the average number of sub-questions generated by the model, and \textbf{Avg. Ret} refers to the average number of sub-questions deemed necessary for retrieval by the model.}
\label{tab:dynamic-generate-full}
\end{table*}

\begin{table*}[h]
\small
\centering
\begin{tabular}{p{0.95\textwidth}}
\hline
You are a powerful multi-hop question generator. Users will provide a chain of Wikidata triplets, and you will help write questions to ask the tail entity from the head entity. \
The format of a wikidata triple is (subject, relation, object). You shouldn’t include bridge entities in generated questions. The questions should only include the head entity. \textbf{All involved relations must be reflected in the question.}
\\
\\
\textbf{\#Example 1} \\
\textbf{Wikidata triplets}:(Four Peaks, mountain range, x1), (x1, located in the administrative territorial entity, x2), (x2, located in the administrative territorial entity, x3), (x3, office held by head of government, x4) \\
\textbf{Generated question}: Who holds the office of the head of government for the administrative entity where the mountain range Four Peaks is located?? \\
\\
\textbf{\#Example 2} \\
\textbf{Wikidata triplets}: (Alena Vostrá, place of birth, x1) \\
\textbf{Generated question}: Where was Alena Vostrá born? \\
\\
\textbf{\#Example 3} \\
\textbf{Wikidata triplets}:  (Anguilla, country, x1), (x1, capital, x2) \\
\textbf{Generated question}:  what is the capital of the country of the Anguilla? \\
\\
\textbf{\#Example 4} \\
\textbf{Wikidata triplets}: (Nazko River, mouth of the watercourse, x1), (x1, mouth of the watercourse, x2), \\
$\langle$x2, country, x3) \\
\textbf{Generated question}: In which country does the Nazko River ultimately discharge its waters?\\
\\
\textbf{\#Example 5} \\
\textbf{Wikidate triplets}: \{Sampled facts\} \\
\textbf{Generated question}:\\
\hline
\end{tabular}
\caption{The prompt used to generate questions is based on sampled facts. Additionally, we include 4 demonstrations showcasing examples ranging from 1-hop to 4-hop reasoning.}
\label{tab:wikidata-templates}
\end{table*}

\begin{table*}[h]
\small
\centering
\begin{tabular}{p{0.95\textwidth}}
\hline
You are a powerful question answering system. Users will provide a question and useful context. The provided context are some wikidata triplets which format is (subject, relation, object). You should answer the question based on the context. The answer should be a single entity or a list of entities. If the answer is a list of entities, you should return the most relevant one. \

Context: \{related documents\} \\

Question: \{question\} \ \\
\hline
\end{tabular}
\caption{The prompt used for question quality inspection provides a given question and its corresponding facts. We aim for the GPT-4o to correctly answer the question based on this information.}
\label{tab:qa-system-format}
\end{table*}
\begin{table}[h]
\centering
\begin{tabular}{p{0.45\textwidth}}
\hline
Below is a question, please answer it directly and keep your answer as short as possible. \
\textbf{Question}: \{question \} \\
\textbf{Answer}: \\
\hline
\end{tabular}
\caption{The prompt designed to guide the model in providing a concise answer directly to the question.}
\label{tab:question-prompt}
\end{table}
\begin{table}
\begin{tabular}{p{0.45\textwidth}}
\hline
Given some related documents: \{retrieved\_documents\}.
This is a  question: \{question\}.
Please answer the question directly. Please keep your answer as short as possible.
\textbf{Answer}: \\
\hline
\end{tabular}
\caption{The prompt instructs the model to provide a concise answer to the question based on the retrieved documents.}
\label{tab:question-prompt}
\end{table}

\begin{table*}[h]
\centering
\begin{tabular}{p{0.95\textwidth}}
\hline
Here is a question: \{question\} \\
To answer this question. You have to three choices now: \\[0.5em]
\textbf{$\langle$choice A$\rangle$} Generate a sub-question. \\
\textbf{$\langle$choice B$\rangle$} Answer the question directly if you are confident to answer it. \\
\textbf{$\langle$choice C$\rangle$} retrieve some document to help you answer the question. \\[0.5em]
If you choose \textbf{$\langle$choice A$\rangle$}, please output: \\
\{\{"choice A": \{\{"sub-question": "your\_sub\_question\_here"\}\}\}\} \\[0.5em]
If you choose \textbf{$\langle$choice B$\rangle$}, please output: \\
\{\{"choice B": \{\{"answer": "your\_answer\_here"\}\}\}\} \\[0.5em]
If you choose \textbf{$\langle$choice C$\rangle$}, please output: \\
\{\{"choice C": retrieval\}\} \\[0.5em]
The final output should be in the form of a JSON string, without any additional content. Please keep your answer as short as possible. \\
Output: \\
\hline
\end{tabular}
\caption{The prompt is used for retrieval tasks, directly generating answers or creating sub-questions for judgment purposes.}
\label{tab:first-step-prompt}
\end{table*}

\begin{table*}[h]
\centering
\begin{tabular}{p{0.95\textwidth}}
\hline
Given a question: \{question\} \\
The subsequent sub-questions: \{sub\_questions\} \\[0.5em]
You have two choices now: \\[0.5em]
\textbf{$\langle$choice A$\rangle$} answer the final sub-question directly. \\
\textbf{$\langle$choice B$\rangle$} retrieve some document to help you answer the question. Just output retrieval as a placeholder. \\[0.5em]
If you choose \textbf{$\langle$choice A$\rangle$}, please output: \\
\{\{"choice A": \{\{"answer": "your\_answer\_here"\}\}\}\} \\[0.5em]
If you choose \textbf{$\langle$choice B$\rangle$}, please output: \\
\{\{"choice B": retrieval\}\} \\[0.5em]
The final output should be in the form of a JSON string, without any additional content. Please keep your answer as short as possible. \\
Output: \\
\hline
\end{tabular}
\caption{The prompt is used for evaluating sub-questions, performing retrieval, or directly generating answers.}
\label{tab:multi-hop-prompt}
\end{table*}

\begin{table*}[h]
\centering
\begin{tabular}{p{0.95\textwidth}}
\hline
Given a main question: \{question\} \\
And sub-question-answer pairs: \{sub\_question\_answer\_pairs\} \\[0.5em]
Please judge if the main question has been finished. You have two choices now: \\[0.5em]
\textbf{$\langle$choice A$\rangle$} The answer can be found in the sub-question-answer pairs. If you choose this choice, please output the final answer. \\
\textbf{$\langle$choice B$\rangle$} The answer cannot be found and a new sub-question needs to be generated. \\[0.5em]
If you choose \textbf{$\langle$choice A$\rangle$}, please output: \\
\{\{"choice A": \{\{"answer": "final\_answer\_here"\}\}\}\} \\[0.5em]
If you choose \textbf{$\langle$choice B$\rangle$}, please output: \\
\{\{"choice B": \{\{"sub-question": "new\_sub-question\_here"\}\}\}\} \\[0.5em]
The final output should be in the form of a JSON string, without any additional content. Please keep your answer as short as possible. \\
Output: \\
\hline
\end{tabular}
\caption{The prompt provides sub-questions and their answers, requiring the model to determine whether the answer to the main question has been found.}
\label{tab:multi-hop-prompt-2-1}
\end{table*}

\begin{table*}[h]
\centering
\begin{tabular}{p{0.95\textwidth}}
\hline
To answer this question, you may need to generate subquestions following these guidelines:\\
Given a main question and optional previous subquestion-answer pairs, you may need to generate subquestions to help answer this main question. Please ensure to only generate subquestions that are relevant to answering the main question. When there are no more subquestions needed, output "finish".\\
\textbf{Input Format}\\
liRequired:\\
- Main Question: [question]\\
Optional:\\
- Previous Subquestion: [subquestion]\\
- Previous Answer: [subanswer]\\
\textbf{Output Format}\\
One of:\\
- Next Subquestion: [new subquestion]\\
- "finish" (when no further subquestions are needed)\\
\textbf{Generation Guidelines}\\
1. Subquestions should:\\
   - Break down complex aspects of the main question\\
   - Follow a logical progression\\
   - Be specific and focused\\
   - Build upon previous answers when available\\
2. Output "finish" when:\\
   - All relevant aspects have been covered\\
   - Further breakdown would not add value\\
   - The question has been fully addressed\\
\textbf{Examples}\\
Example 1:\\
Input:\\
- Main Question: "What is the location of the headquarters of the institution where Percival Lowell was educated?"\\
- Previous Subquestion: "Where did Percival Lowell receive his education?"\\
- Previous Answer: "Harvard University."\\
Output:\\
- Next Subquestion: "Where is the headquarters of Harvard University?"\\
Example 2:\\
Input:\\
- Main Question: "What is the capital of France?"\\
Output:\\
- "finish"\\
Main Question: \{question\}\\
\{previous\_subquestion\_answer\_pairs\}\\
Output:\\\hline
\end{tabular}\caption{The prompt instructs the model to decompose the main question further by generating sub-questions based on the previous response history.}
    \label{tab:my_label}
\end{table*}

\begin{table}[]
    \centering
\begin{tabular}{p{0.45\textwidth}}
\hline
Based on the main question and all subquestion-answer pairs, please provide a comprehensive final answer. Please keep your answer as short as possible.\\
Main Question: \{main\_question\}\\
Previous Subquestions and Answers:\\
\{history\_str\}\\
Final Answer:\\\hline
\end{tabular}    \caption{The prompt instructs the model to summarize and generate the answer to the main question based on the sub-questions and their answers.}
    \label{tab:my_label}
\end{table}

\begin{table}[]
    \centering
\begin{tabular}{p{0.45\textwidth}}
\hline
Answer the following question based on your internal knowledge with one or few words.\\
Add a confidence indicator after your answer:
- "certain" if you are completely confident in the accuracy
- "uncertain" if you have any doubts\\
\textbf{Input Format}\\
Input:\\
- Question: [question]\\
\textbf{Output Format}\\
Output: \\
- Answer: [brief answer]\\
- Confidence: [certain/uncertain]\\
Question: \{question\}\\
Output:\\\hline
\end{tabular}    \caption{The prompt requires the model to output a confidence score for the generated sub-questions, which will be used to determine whether retrieval is necessary.}
    \label{confidence_task_prompt}
\end{table}


\begin{figure*}
    \centering
    \includegraphics[width=1.0\linewidth]{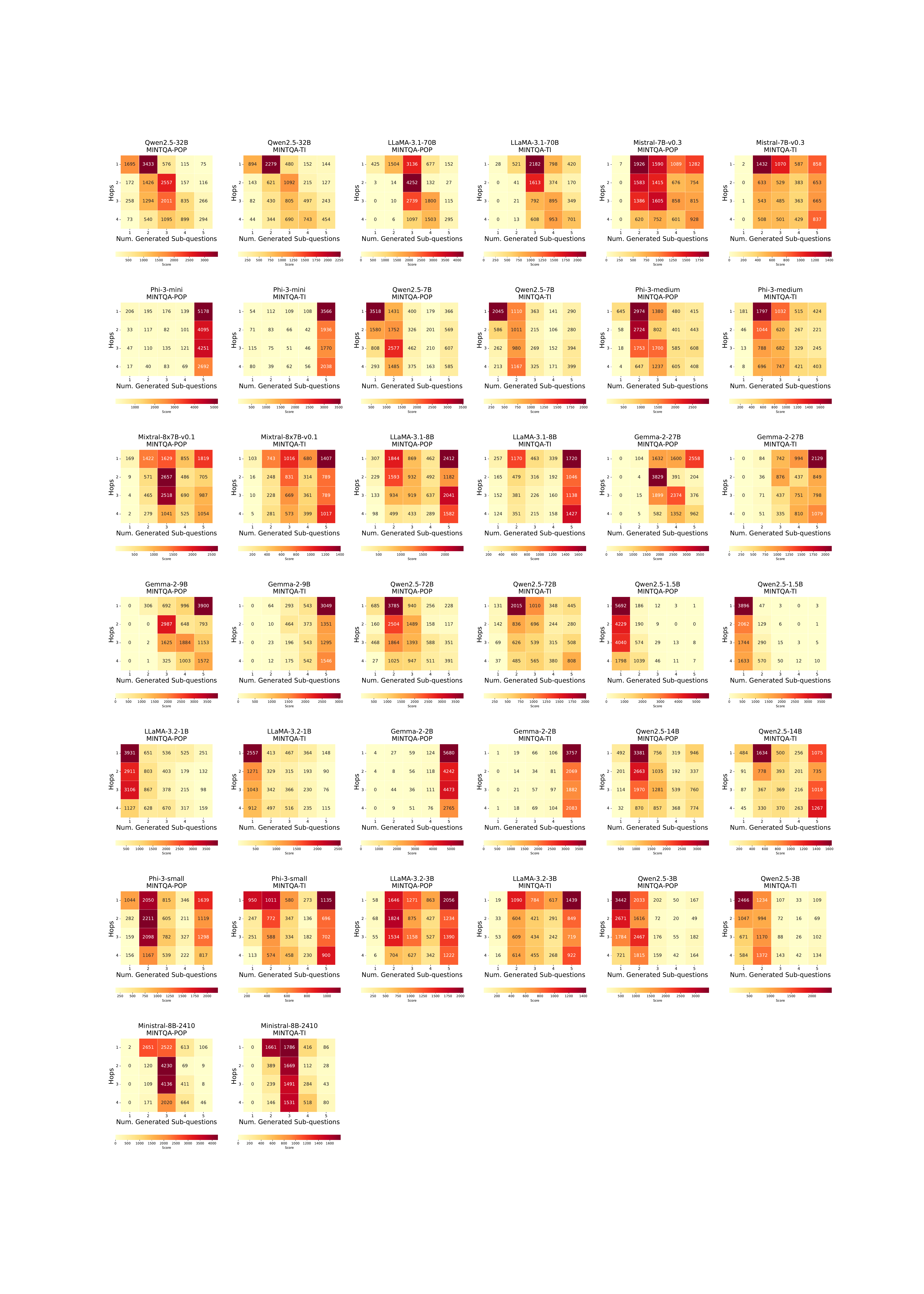}
    \caption{The confusion matrix of the number of sub-questions generated by the large language models for  main questions categorized by  hops in the setting of purely generating sub-questions.}
    \label{fig:matrix}
\end{figure*}



\begin{table*}[ht]\centering\scriptsize\begin{tabular}{p{15cm}}\hline
\textbf{Triplets:} [[Pigeon Bay Domain, country, New Zealand]]\\
\textbf{Main Question:} In which country is Pigeon Bay Domain located?\\
\textbf{Main Answer:} New Zealand\\
\textbf{Type:} New\\
\hline
\textbf{Triplets:} [[Eveline Hoffmann, place of detention, Theresienstadt Ghetto]]\\
\textbf{Main Question:} Where was Eveline Hoffmann detained?\\
\textbf{Main Answer:} Theresienstadt Ghetto\\
\textbf{Type:} Old\\
\hline
\end{tabular}\caption{One-hop question-answer pairs and their corresponding types in \textsc{MINTQA-ti}.}
\label{first_case}
\end{table*}
\begin{table*}[ht]\centering\scriptsize\begin{tabular}{p{15cm}}\hline
\textbf{Triplets:} [[Scram Kitty and his Buddy on Rails, publisher, Dakko Dakko], [Dakko Dakko, industry, video game industry]]\\
\textbf{Main Question:} In which industry does the publisher of Scram Kitty and his Buddy on Rails operate?\\
\textbf{Main Answer:} video game industry\\
\textbf{Subquestion pairs:}\\
\textbf{Sub-question} 0: Who is the publisher of Scram Kitty and his Buddy on Rails? \textbf{Sub-answer} 0: Dakko Dakko. \textbf{Type}: New\\
\textbf{Sub-question} 1: In which industry does Dakko Dakko operate? \textbf{Sub-answer} 1: video game industry. \textbf{Type}: New\\
\hline
\textbf{Triplets:} [[CineKink NYC, location, New York City], [New York City, capital of, United States of America]]\\
\textbf{Main Question:} CineKink NYC is located in the city that is the capital of which entity?\\
\textbf{Main Answer:} United States of America\\
\textbf{Subquestion pairs:}\\
\textbf{Sub-question} 0: Where is CineKink NYC located? \textbf{Sub-answer} 0: New York City. \textbf{Type}: New\\
\textbf{Sub-question} 1: What entity has New York City as its capital? \textbf{Sub-answer} 1: United States of America. \textbf{Type}: Old\\
\hline
\textbf{Triplets:} [[Sanna Aunesluoma, residence, Espoo], [Espoo, member of, Union of the Baltic Cities]]\\
\textbf{Main Question:} Which organization or group is the residence of Sanna Aunesluoma a member of?\\
\textbf{Main Answer:} Union of the Baltic Cities\\
\textbf{Subquestion pairs:}\\
\textbf{Sub-question} 0: Where does Sanna Aunesluoma reside? \textbf{Sub-answer} 0: Espoo. \textbf{Type}: Old\\
\textbf{Sub-question} 1: Of which entity is Espoo a member? \textbf{Sub-answer} 1: Union of the Baltic Cities. \textbf{Type}: New\\
\hline
\textbf{Triplets:} [[Horst Hoffmann, country of citizenship, German Democratic Republic], [German Democratic Republic, legislative body, Volkskammer]]\\
\textbf{Main Question:} What is the legislative body of the country where Horst Hoffmann holds citizenship?\\
\textbf{Main Answer:} Volkskammer\\
\textbf{Subquestion pairs:}\\
\textbf{Sub-question} 0: What is the country of citizenship of Horst Hoffmann? \textbf{Sub-answer} 0: German Democratic Republic. \textbf{Type}: Old\\
\textbf{Sub-question} 1: What is the legislative body of the German Democratic Republic? \textbf{Sub-answer} 1: Volkskammer. \textbf{Type}: Old\\
\hline
\end{tabular}\caption{Two-hop question-answer pairs and their corresponding types in \textsc{MINTQA-ti}.}\end{table*}

\begin{table*}[ht]\centering\scriptsize\begin{tabular}{p{15cm}}\hline
\textbf{Triplets:} [[Systems and methods for mesh augmentation and prevention of incisional hernia, owned by, The Trustees of the University of Pennsylvania], [The Trustees of the University of Pennsylvania, headquarters location, Philadelphia], [Philadelphia, member of, Organization of World Heritage Cities]]\\
\textbf{Main Question:} Of which entity is the headquarters location of the owner of the "Systems and methods for mesh augmentation and prevention of incisional hernia" a member?\\
\textbf{Main Answer:} Organization of World Heritage Cities\\
\textbf{Subquestion pairs:}\\
\textbf{Sub-question} 0: Who owns the patent for Systems and methods for mesh augmentation and prevention of incisional hernia? \textbf{Sub-answer} 0: The Trustees of the University of Pennsylvania. \textbf{Type}: New\\
\textbf{Sub-question} 1: Where is the headquarters of The Trustees of the University of Pennsylvania located? \textbf{Sub-answer} 1: Philadelphia. \textbf{Type}: New\\
\textbf{Sub-question} 2: What is Philadelphia a member of? \textbf{Sub-answer} 2: Organization of World Heritage Cities. \textbf{Type}: New\\
\hline
\textbf{Triplets:} [[De grote Gwen en Geraldine show, nominated for, Dutch Podcast Award for Chatcast Vermaak], [Dutch Podcast Award for Chatcast Vermaak, country, Netherlands], [Netherlands, language used, Dutch]]\\
\textbf{Main Question:} What is the language used in the country for which "De grote Gwen en Geraldine show" was nominated?\\
\textbf{Main Answer:} Dutch\\
\textbf{Subquestion pairs:}\\
\textbf{Sub-question} 0: For what award was "De grote Gwen en Geraldine show" nominated? \textbf{Sub-answer} 0: Dutch Podcast Award for Chatcast Vermaak. \textbf{Type}: New\\
\textbf{Sub-question} 1: In which country is the Dutch Podcast Award for Chatcast Vermaak given? \textbf{Sub-answer} 1: Netherlands. \textbf{Type}: New\\
\textbf{Sub-question} 2: What language is used in the Netherlands? \textbf{Sub-answer} 2: Dutch. \textbf{Type}: Old\\
\hline
\textbf{Triplets:} [[Gathering to Celebrate Old Age, creator, Tomioka Tessai], [Tomioka Tessai, location, Tokyo National Museum], [Tokyo National Museum, member of, Japan Consortium for Open Access Repository]]\\
\textbf{Main Question:} Which organization or group is the location associated with the creator of "Gathering to Celebrate Old Age" a member of?\\
\textbf{Main Answer:} Japan Consortium for Open Access Repository\\
\textbf{Subquestion pairs:}\\
\textbf{Sub-question} 0: Who is the creator of Gathering to Celebrate Old Age? \textbf{Sub-answer} 0: Tomioka Tessai. \textbf{Type}: New\\
\textbf{Sub-question} 1: Where is Tomioka Tessai located? \textbf{Sub-answer} 1: Tokyo National Museum. \textbf{Type}: Old\\
\textbf{Sub-question} 2: What organization or association is the Tokyo National Museum a member of? \textbf{Sub-answer} 2: Japan Consortium for Open Access Repository. \textbf{Type}: New\\
\hline
\textbf{Triplets:} [[The Woman Who Cooked Her Husband, author, Debbie Isitt], [Debbie Isitt, country of citizenship, United Kingdom], [United Kingdom, continent, Europe]]\\
\textbf{Main Question:} On which continent does the author of "The Woman Who Cooked Her Husband" hold citizenship?\\
\textbf{Main Answer:} Europe\\
\textbf{Subquestion pairs:}\\
\textbf{Sub-question} 0: Who is the author of "The Woman Who Cooked Her Husband"? \textbf{Sub-answer} 0: Debbie Isitt. \textbf{Type}: New\\
\textbf{Sub-question} 1: What is the country of citizenship of Debbie Isitt? \textbf{Sub-answer} 1: United Kingdom. \textbf{Type}: Old\\
\textbf{Sub-question} 2: On which continent is the United Kingdom located? \textbf{Sub-answer} 2: Europe. \textbf{Type}: Old\\
\hline
\textbf{Triplets:} [[Mubarak Shah, religion or worldview, Islam], [Islam, item operated, Qalab], [Qalab, cause of death, Ajal]]\\
\textbf{Main Question:} What was the cause of death for the operator of the religion or worldview followed by Mubarak Shah?\\
\textbf{Main Answer:} Ajal\\
\textbf{Subquestion pairs:}\\
\textbf{Sub-question} 0: What is the religion or worldview of Mubarak Shah? \textbf{Sub-answer} 0: Islam. \textbf{Type}: Old\\
\textbf{Sub-question} 1: What item is operated by Islam? \textbf{Sub-answer} 1: Qalab. \textbf{Type}: New\\
\textbf{Sub-question} 2: What was the cause of death for Qalab? \textbf{Sub-answer} 2: Ajal. \textbf{Type}: New\\
\hline
\textbf{Triplets:} [[Felipe Borrego Estrada, place of birth, Zacatecas], [Zacatecas, member of, Organization of World Heritage Cities], [Organization of World Heritage Cities, headquarters location, Quebec City]]\\
\textbf{Main Question:} Where is the headquarters of the entity that the birthplace of Felipe Borrego Estrada is a member of?\\
\textbf{Main Answer:} Quebec City\\
\textbf{Subquestion pairs:}\\
\textbf{Sub-question} 0: Where was Felipe Borrego Estrada born? \textbf{Sub-answer} 0: Zacatecas. \textbf{Type}: Old\\
\textbf{Sub-question} 1: Of which organization is Zacatecas a member? \textbf{Sub-answer} 1: Organization of World Heritage Cities. \textbf{Type}: New\\
\textbf{Sub-question} 2: Where is the headquarters of the Organization of World Heritage Cities located? \textbf{Sub-answer} 2: Quebec City. \textbf{Type}: Old\\
\hline
\textbf{Triplets:} [[Hykjeberget, operator, Dalarna County Administrative Board], [Dalarna County Administrative Board, headquarters location, Falun], [Falun, twinned administrative body, Hamina]]\\
\textbf{Main Question:} What administrative body is twinned with the location of the headquarters of the operator of Hykjeberget?\\
\textbf{Main Answer:} Hamina\\
\textbf{Subquestion pairs:}\\
\textbf{Sub-question} 0: Who operates Hykjeberget? \textbf{Sub-answer} 0: Dalarna County Administrative Board. \textbf{Type}: Old\\
\textbf{Sub-question} 1: Where is the headquarters of the Dalarna County Administrative Board located? \textbf{Sub-answer} 1: Falun. \textbf{Type}: Old\\
\textbf{Sub-question} 2: Which administrative body is twinned with Falun? \textbf{Sub-answer} 2: Hamina. \textbf{Type}: New\\
\hline
\textbf{Triplets:} [[University of California Italian Studies Multicampus Research Group, country, United States of America], [United States of America, highest point, Denali], [Denali, mountain range, Alaska Range]]\\
\textbf{Main Question:} What is the mountain range that contains the highest point in the country where the University of California Italian Studies Multicampus Research Group is located?\\
\textbf{Main Answer:} Alaska Range\\
\textbf{Subquestion pairs:}\\
\textbf{Sub-question} 0: In which country is the University of California Italian Studies Multicampus Research Group located? \textbf{Sub-answer} 0: United States of America. \textbf{Type}: Old\\
\textbf{Sub-question} 1: What is the highest point in the United States of America? \textbf{Sub-answer} 1: Denali. \textbf{Type}: Old\\
\textbf{Sub-question} 2: In which mountain range is Denali located? \textbf{Sub-answer} 2: Alaska Range. \textbf{Type}: Old\\
\hline
\end{tabular}\caption{Three-hop question-answer pairs and their corresponding types in \textsc{MINTQA-ti}.}\end{table*}

\begin{table*}[ht]\centering\scriptsize\begin{tabular}{p{15cm}}\hline
\textbf{Triplets:} [[Patricia Florence Suthers, sibling, Elaine Suthers], [Elaine Suthers, mother, Elsie Suthers], [Elsie Suthers, country of citizenship, United Kingdom], [United Kingdom, highest point, Ben Nevis]]\\
\textbf{Main Question:} What is the highest point in the country where the mother of Patricia Florence Suthers' sibling is a citizen?\\
\textbf{Main Answer:} Ben Nevis\\
\textbf{Subquestion pairs:}\\
\textbf{Sub-question} 0: Who is the sibling of Patricia Florence Suthers? \textbf{Sub-answer} 0: Elaine Suthers. \textbf{Type}: New\\
\textbf{Sub-question} 1: Who is the mother of Elaine Suthers? \textbf{Sub-answer} 1: Elsie Suthers. \textbf{Type}: New\\
\textbf{Sub-question} 2: Which country is Elsie Suthers a citizen of? \textbf{Sub-answer} 2: United Kingdom. \textbf{Type}: New\\
\textbf{Sub-question} 3: What is the highest point in the United Kingdom? \textbf{Sub-answer} 3: Ben Nevis. \textbf{Type}: Old\\
\hline
\textbf{Triplets:} [[Patricia Florence Suthers, mother, Elsie Suthers], [Elsie Suthers, spouse, Robert Suthers], [Robert Suthers, relative, Miriam Farid], [Miriam Farid, country of citizenship, United Kingdom]]\\
\textbf{Main Question:} What is the country of citizenship of the relative of Patricia Florence Suthers' mother's spouse?\\
\textbf{Main Answer:} United Kingdom\\
\textbf{Subquestion pairs:}\\
\textbf{Sub-question} 0: Who is the mother of Patricia Florence Suthers? \textbf{Sub-answer} 0: Elsie Suthers. \textbf{Type}: New\\
\textbf{Sub-question} 1: Who is the spouse of Elsie Suthers? \textbf{Sub-answer} 1: Robert Suthers. \textbf{Type}: New\\
\textbf{Sub-question} 2: Who is a relative of Robert Suthers? \textbf{Sub-answer} 2: Miriam Farid. \textbf{Type}: New\\
\textbf{Sub-question} 3: Which country is Miriam Farid a citizen of? \textbf{Sub-answer} 3: United Kingdom. \textbf{Type}: New\\
\hline
\textbf{Triplets:} [[May Hnin Aw Kanya, mother, May Hnin Htapi], [May Hnin Htapi, father, Loethai], [Loethai, child, Lithai], [Lithai, notable work, Traibhumikatha]]\\
\textbf{Main Question:} What is the notable work of the child of the father of the mother of May Hnin Aw Kanya?\\
\textbf{Main Answer:} Traibhumikatha\\
\textbf{Subquestion pairs:}\\
\textbf{Sub-question} 0: Who is the mother of May Hnin Aw Kanya? \textbf{Sub-answer} 0: May Hnin Htapi. \textbf{Type}: New\\
\textbf{Sub-question} 1: Who is May Hnin Htapi's father? \textbf{Sub-answer} 1: Loethai. \textbf{Type}: New\\
\textbf{Sub-question} 2: Who is the child of Loethai? \textbf{Sub-answer} 2: Lithai. \textbf{Type}: Old\\
\textbf{Sub-question} 3: What is a notable work created by Lithai? \textbf{Sub-answer} 3: Traibhumikatha. \textbf{Type}: New\\
\hline
\textbf{Triplets:} [[SEOlytics, parent organization, Sistrix], [Sistrix, country, Germany], [Germany, continent, Europe], [Europe, shares border with, Asia]]\\
\textbf{Main Question:} Which continent shares a border with the continent where the country of SEOlytics' parent organization is located?\\
\textbf{Main Answer:} Asia\\
\textbf{Subquestion pairs:}\\
\textbf{Sub-question} 0: What is the parent organization of SEOlytics? \textbf{Sub-answer} 0: Sistrix. \textbf{Type}: New\\
\textbf{Sub-question} 1: In which country is Sistrix located? \textbf{Sub-answer} 1: Germany. \textbf{Type}: New\\
\textbf{Sub-question} 2: On which continent is Germany located? \textbf{Sub-answer} 2: Europe. \textbf{Type}: Old\\
\textbf{Sub-question} 3: Which continent shares a border with Europe? \textbf{Sub-answer} 3: Asia. \textbf{Type}: Old\\
\hline
\textbf{Triplets:} [[Sri Dhamasokaraj, relative, Saileuthai], [Saileuthai, father, Lithai], [Lithai, sibling, May Hnin Htapi], [May Hnin Htapi, place of death, Mottama]]\\
\textbf{Main Question:} Where did the sibling of the father of Sri Dhamasokaraj pass away?\\
\textbf{Main Answer:} Mottama\\
\textbf{Subquestion pairs:}\\
\textbf{Sub-question} 0: Who is a relative of Sri Dhamasokaraj? \textbf{Sub-answer} 0: Saileuthai. \textbf{Type}: New\\
\textbf{Sub-question} 1: Who was the father of Saileuthai? \textbf{Sub-answer} 1: Lithai. \textbf{Type}: Old\\
\textbf{Sub-question} 2: Who is Lithai's sibling? \textbf{Sub-answer} 2: May Hnin Htapi. \textbf{Type}: New\\
\textbf{Sub-question} 3: Where did May Hnin Htapi die? \textbf{Sub-answer} 3: Mottama. \textbf{Type}: New\\
\hline
\textbf{Triplets:} [[Frank Gailor, educated at, New College], [New College, founded by, William of Wykeham], [William of Wykeham, country of citizenship, Kingdom of England], [Kingdom of England, replaced by, Kingdom of Great Britain]]\\
\textbf{Main Question:} Which entity replaced the country of citizenship of the founder of the institution where Frank Gailor was educated?\\
\textbf{Main Answer:} Kingdom of Great Britain\\
\textbf{Subquestion pairs:}\\
\textbf{Sub-question} 0: Where was Frank Gailor educated? \textbf{Sub-answer} 0: New College. \textbf{Type}: New\\
\textbf{Sub-question} 1: Who founded New College? \textbf{Sub-answer} 1: William of Wykeham. \textbf{Type}: Old\\
\textbf{Sub-question} 2: Which country was William of Wykeham a citizen of? \textbf{Sub-answer} 2: Kingdom of England. \textbf{Type}: New\\
\textbf{Sub-question} 3: What entity replaced the Kingdom of England? \textbf{Sub-answer} 3: Kingdom of Great Britain. \textbf{Type}: Old\\
\hline 
\textbf{Triplets:} [[The Life You Can Save, author, Peter Singer], [Peter Singer, mother, Cora Singer], [Cora Singer, father, David Ernst Oppenheim], [David Ernst Oppenheim, academic degree, doctorate]]\\
\textbf{Main Question:} What academic degree does the father of the author of "The Life You Can Save" hold?\\
\textbf{Main Answer:} doctorate\\
\textbf{Subquestion pairs:}\\
\textbf{Sub-question} 0: Who is the author of "The Life You Can Save"? \textbf{Sub-answer} 0: Peter Singer. \textbf{Type}: Old\\
\textbf{Sub-question} 1: Who is Peter Singer's mother? \textbf{Sub-answer} 1: Cora Singer. \textbf{Type}: New\\
\textbf{Sub-question} 2: Who is the father of Cora Singer? \textbf{Sub-answer} 2: David Ernst Oppenheim. \textbf{Type}: New\\
\textbf{Sub-question} 3: What academic degree does David Ernst Oppenheim hold? \textbf{Sub-answer} 3: doctorate. \textbf{Type}: Old\\
\hline
\textbf{Triplets:} [[Geoffrey Howe, creator, June Mendoza], [June Mendoza, place of birth, Melbourne], [Melbourne, located in or next to body of water, Yarra River], [Yarra River, continent, Australian continent]]\\
\textbf{Main Question:} On which continent is the body of water located next to the place where the creator Geoffrey Howe was born?\\
\textbf{Main Answer:} Australian continent\\
\textbf{Subquestion pairs:}\\
\textbf{Sub-question} 0: What did Geoffrey Howe create? \textbf{Sub-answer} 0: June Mendoza. \textbf{Type}: Old\\
\textbf{Sub-question} 1: Where was June Mendoza born? \textbf{Sub-answer} 1: Melbourne. \textbf{Type}: New\\
\textbf{Sub-question} 2: Which body of water is Melbourne located near? \textbf{Sub-answer} 2: Yarra River. \textbf{Type}: Old\\
\textbf{Sub-question} 3: On which continent is the Yarra River located? \textbf{Sub-answer} 3: Australian continent. \textbf{Type}: New\\
\hline 
\end{tabular}\caption{Four-hop question-answer pairs and their corresponding types in \textsc{MINTQA-ti} (part 1).}\end{table*}

\begin{table*}[ht]\centering\scriptsize\begin{tabular}{p{15cm}}\hline
\textbf{Triplets:} [[Descenso a los fascismos, place of publication, Barcelona], [Barcelona, member of, Creative Cities Network], [Creative Cities Network, operator, UNESCO], [UNESCO, operating area, worldwide]]\\
\textbf{Main Question:} In what area does the operator of the organization that includes the place where "Descenso a los fascismos" was published operate?\\
\textbf{Main Answer:} worldwide\\
\textbf{Subquestion pairs:}\\
\textbf{Sub-question} 0: Where was "Descenso a los fascismos" published? \textbf{Sub-answer} 0: Barcelona. \textbf{Type}: New\\
\textbf{Sub-question} 1: What organization or group is Barcelona a member of? \textbf{Sub-answer} 1: Creative Cities Network. \textbf{Type}: Old\\
\textbf{Sub-question} 2: Who operates the Creative Cities Network? \textbf{Sub-answer} 2: UNESCO. \textbf{Type}: Old\\
\textbf{Sub-question} 3: What is the operating area of UNESCO? \textbf{Sub-answer} 3: worldwide. \textbf{Type}: New\\
\hline
\textbf{Triplets:} [[Monument to Terenzio Mamiani, commemorates, Terenzio, Count Mamiani della Rovere], [Terenzio, Count Mamiani della Rovere, award received, Order of the Redeemer], [Order of the Redeemer, founded by, Otto of Greece], [Otto of Greece, spouse, Amalia of Oldenburg]]\\
\textbf{Main Question:} Who is the spouse of the founder of the award received by the person commemorated by the Monument to Terenzio Mamiani?\\
\textbf{Main Answer:} Amalia of Oldenburg\\
\textbf{Subquestion pairs:}\\
\textbf{Sub-question} 0: Who is commemorated by the Monument to Terenzio Mamiani? \textbf{Sub-answer} 0: Terenzio, Count Mamiani della Rovere. \textbf{Type}: New\\
\textbf{Sub-question} 1: What award did Terenzio, Count Mamiani della Rovere receive? \textbf{Sub-answer} 1: Order of the Redeemer. \textbf{Type}: Old\\
\textbf{Sub-question} 2: Who founded the Order of the Redeemer? \textbf{Sub-answer} 2: Otto of Greece. \textbf{Type}: Old\\
\textbf{Sub-question} 3: Who was the spouse of Otto of Greece? \textbf{Sub-answer} 3: Amalia of Oldenburg. \textbf{Type}: Old\\
\hline
\textbf{Triplets:} [[Tansen, religion or worldview, Islam], [Islam, item operated, Qalab], [Qalab, cause of death, Ajal], [Ajal, location, treasures of God in Islam]]\\
\textbf{Main Question:} Where did the cause of death of the religious figure associated with Tansen occur?\\
\textbf{Main Answer:} treasures of God in Islam\\
\textbf{Subquestion pairs:}\\
\textbf{Sub-question} 0: What is the religion or worldview associated with Tansen? \textbf{Sub-answer} 0: Islam. \textbf{Type}: Old\\
\textbf{Sub-question} 1: What item is operated by Islam? \textbf{Sub-answer} 1: Qalab. \textbf{Type}: New\\
\textbf{Sub-question} 2: What was the cause of death for Qalab? \textbf{Sub-answer} 2: Ajal. \textbf{Type}: New\\
\textbf{Sub-question} 3: Where is Ajal located? \textbf{Sub-answer} 3: treasures of God in Islam. \textbf{Type}: New\\
\hline
\textbf{Triplets:} [[Irma Stern, place of birth, Bratislava], [Bratislava, member of, League of Historical Cities], [League of Historical Cities, headquarters location, Kyoto], [Kyoto, highest point, Mount Minako]]\\
\textbf{Main Question:} What is the highest point of the location where the headquarters of the entity that includes the birthplace of Irma Stern is situated?\\
\textbf{Main Answer:} Mount Minako\\
\textbf{Subquestion pairs:}\\
\textbf{Sub-question} 0: Where was Irma Stern born? \textbf{Sub-answer} 0: Bratislava. \textbf{Type}: Old\\
\textbf{Sub-question} 1: Of which organization is Bratislava a member? \textbf{Sub-answer} 1: League of Historical Cities. \textbf{Type}: New\\
\textbf{Sub-question} 2: Where is the headquarters of the League of Historical Cities located? \textbf{Sub-answer} 2: Kyoto. \textbf{Type}: Old\\
\textbf{Sub-question} 3: What is the highest point in Kyoto? \textbf{Sub-answer} 3: Mount Minako. \textbf{Type}: Old\\
\hline
\textbf{Triplets:} [[Andrew Cogglesby, present in work, Evan Harrington], [Evan Harrington, author, George Meredith], [George Meredith, spouse, Mary Meredith], [Mary Meredith, cause of death, kidney failure]]\\
\textbf{Main Question:} What was the cause of death of the spouse of the author who created the work featuring Andrew Cogglesby?\\
\textbf{Main Answer:} kidney failure\\
\textbf{Subquestion pairs:}\\
\textbf{Sub-question} 0: In which work does Andrew Cogglesby appear? \textbf{Sub-answer} 0: Evan Harrington. \textbf{Type}: Old\\
\textbf{Sub-question} 1: Who is the author of "Evan Harrington"? \textbf{Sub-answer} 1: George Meredith. \textbf{Type}: Old\\
\textbf{Sub-question} 2: Who is the spouse of George Meredith? \textbf{Sub-answer} 2: Mary Meredith. \textbf{Type}: New\\
\textbf{Sub-question} 3: What was the cause of death of Mary Meredith? \textbf{Sub-answer} 3: kidney failure. \textbf{Type}: New\\
\hline
\textbf{Triplets:} [[Federico Cocozza, employer, Curie Institute], [Curie Institute, founded by, Marie Curie], [Marie Curie, ethnic group, Poles], [Poles, language used, Church Slavonic]]\\
\textbf{Main Question:} What language is used by the ethnic group of the founder of Federico Cocozza's employer?\\
\textbf{Main Answer:} Church Slavonic\\
\textbf{Subquestion pairs:}\\
\textbf{Sub-question} 0: Who employs Federico Cocozza? \textbf{Sub-answer} 0: Curie Institute. \textbf{Type}: Old\\
\textbf{Sub-question} 1: Who founded the Curie Institute? \textbf{Sub-answer} 1: Marie Curie. \textbf{Type}: Old\\
\textbf{Sub-question} 2: What is the ethnic group of Marie Curie? \textbf{Sub-answer} 2: Poles. \textbf{Type}: New\\
\textbf{Sub-question} 3: Which language is used by Poles? \textbf{Sub-answer} 3: Church Slavonic. \textbf{Type}: Old\\
\hline
\textbf{Triplets:} [[Devespresso Games, headquarters location, Seoul], [Seoul, member of, Creative Cities Network], [Creative Cities Network, operator, UNESCO], [UNESCO, operating area, worldwide]]\\
\textbf{Main Question:} What is the operating area of the operator of the member organization where Devespresso Games’ headquarters is located?\\
\textbf{Main Answer:} worldwide\\
\textbf{Subquestion pairs:}\\
\textbf{Sub-question} 0: Where is the headquarters of Devespresso Games located? \textbf{Sub-answer} 0: Seoul. \textbf{Type}: Old\\
\textbf{Sub-question} 1: Of which organization is Seoul a member? \textbf{Sub-answer} 1: Creative Cities Network. \textbf{Type}: Old\\
\textbf{Sub-question} 2: Who operates the Creative Cities Network? \textbf{Sub-answer} 2: UNESCO. \textbf{Type}: Old\\
\textbf{Sub-question} 3: What is the operating area of UNESCO? \textbf{Sub-answer} 3: worldwide. \textbf{Type}: New\\
\hline
\textbf{Triplets:} [[Sonetto I, author, Vittorio Alfieri], [Vittorio Alfieri, place of death, Florence], [Florence, present in work, Civilization V], [Civilization V, developer, Firaxis Games]]\\
\textbf{Main Question:} Who is the developer of the work where the place of death of the author of Sonetto I is present?\\
\textbf{Main Answer:} Firaxis Games\\
\textbf{Subquestion pairs:}\\
\textbf{Sub-question} 0: Who is the author of Sonetto I? \textbf{Sub-answer} 0: Vittorio Alfieri. \textbf{Type}: Old\\
\textbf{Sub-question} 1: Where did Vittorio Alfieri die? \textbf{Sub-answer} 1: Florence. \textbf{Type}: Old\\
\textbf{Sub-question} 2: In which work is Florence present? \textbf{Sub-answer} 2: Civilization V. \textbf{Type}: Old\\
\textbf{Sub-question} 3: Who developed Civilization V? \textbf{Sub-answer} 3: Firaxis Games. \textbf{Type}: Old\\
\hline
\end{tabular}\caption{Four-hop question-answer pairs and their corresponding types in \textsc{MINTQA-ti} (part 2)..}\end{table*}
\begin{table*}[ht]\centering\scriptsize\begin{tabular}{p{15cm}}\hline
\textbf{Triplets:} [[Papanasam taluk, country, India]]\\
\textbf{Main Question:} In which country is Papanasam taluk located?\\
\textbf{Main Answer:} India\\
\textbf{Type:} Popular\\
\hline
\textbf{Triplets:} [[Jerod Swallow, sports discipline competed in, ice dance]]\\
\textbf{Main Question:} In which sports discipline does Jerod Swallow compete?\\
\textbf{Main Answer:} ice dance\\
\textbf{Type:} Unpopular\\
\hline
\end{tabular}\caption{One-hop question-answer pairs and their corresponding types in \textsc{MINTQA-pop}.}\end{table*}
\begin{table*}[ht]\centering\scriptsize\begin{tabular}{p{15cm}}\hline
\textbf{Triplets:} [[Gmina Szypliszki, country, Poland], [Poland, capital, Warsaw]]\\
\textbf{Main Question:} What is the capital of the country where Gmina Szypliszki is located?\\
\textbf{Main Answer:} Warsaw\\
\textbf{Subquestion pairs:}\\
\textbf{Sub-question} 0: In which country is Gmina Szypliszki located? \textbf{Sub-answer} 0: Poland. \textbf{Type}: Popular\\
\textbf{Sub-question} 1: What is the capital of Poland? \textbf{Sub-answer} 1: Warsaw. \textbf{Type}: Popular\\
\hline
\textbf{Triplets:} [[Canary Islands, country, Spain], [Spain, legislative body, Cortes Generales]]\\
\textbf{Main Question:} What is the legislative body of the country to which the Canary Islands belong?\\
\textbf{Main Answer:} Cortes Generales\\
\textbf{Subquestion pairs:}\\
\textbf{Sub-question} 0: Which country are the Canary Islands part of? \textbf{Sub-answer} 0: Spain. \textbf{Type}: Popular\\
\textbf{Sub-question} 1: What is the legislative body of Spain? \textbf{Sub-answer} 1: Cortes Generales. \textbf{Type}: Unpopular\\
\hline
\textbf{Triplets:} [[Pabna Cadet College, country, Bangladesh], [Bangladesh, capital, Dhaka]]\\
\textbf{Main Question:} What is the capital of the country where Pabna Cadet College is located?\\
\textbf{Main Answer:} Dhaka\\
\textbf{Subquestion pairs:}\\
\textbf{Sub-question} 0: In which country is Pabna Cadet College located? \textbf{Sub-answer} 0: Bangladesh. \textbf{Type}: Unpopular\\
\textbf{Sub-question} 1: What is the capital of Bangladesh? \textbf{Sub-answer} 1: Dhaka. \textbf{Type}: Popular\\
\hline
\textbf{Triplets:} [[Brackendale Eagles Provincial Park, country, Canada], [Canada, highest point, Mount Logan]]\\
\textbf{Main Question:} What is the highest point in the country where Brackendale Eagles Provincial Park is located?\\
\textbf{Main Answer:} Mount Logan\\
\textbf{Subquestion pairs:}\\
\textbf{Sub-question} 0: In which country is Brackendale Eagles Provincial Park located? \textbf{Sub-answer} 0: Canada. \textbf{Type}: Unpopular\\
\textbf{Sub-question} 1: What is the highest point in Canada? \textbf{Sub-answer} 1: Mount Logan. \textbf{Type}: Unpopular\\
\hline
\end{tabular}\caption{Two-hop question-answer pairs and their corresponding types in \textsc{MINTQA-pop}.}\end{table*}
\begin{table*}[ht]\centering\scriptsize\begin{tabular}{p{15cm}}\hline
\textbf{Triplets:} [[Cuzco Department, country, Peru], [Peru, capital, Lima], [Lima, located in or next to body of water, Rímac River]]\\
\textbf{Main Question:} Which body of water is located in or next to the capital of the country where the Cuzco Department is found?\\
\textbf{Main Answer:} Rímac River\\
\textbf{Subquestion pairs:}\\
\textbf{Sub-question} 0: In which country is the Cuzco Department located? \textbf{Sub-answer} 0: Peru. \textbf{Type}: Popular\\
\textbf{Sub-question} 1: What is the capital of Peru? \textbf{Sub-answer} 1: Lima. \textbf{Type}: Popular\\
\textbf{Sub-question} 2: Which body of water is Lima located next to? \textbf{Sub-answer} 2: Rímac River. \textbf{Type}: Unpopular\\
\hline
\textbf{Triplets:} [[Kirkovo Municipality, country, Bulgaria], [Bulgaria, highest point, Musala], [Musala, mountain range, Rila]]\\
\textbf{Main Question:} Which mountain range includes the highest point in the country of Kirkovo Municipality?\\
\textbf{Main Answer:} Rila\\
\textbf{Subquestion pairs:}\\
\textbf{Sub-question} 0: Which country is Kirkovo Municipality located in? \textbf{Sub-answer} 0: Bulgaria. \textbf{Type}: Popular\\
\textbf{Sub-question} 1: What is the highest point in Bulgaria? \textbf{Sub-answer} 1: Musala. \textbf{Type}: Unpopular\\
\textbf{Sub-question} 2: In which mountain range is Musala located? \textbf{Sub-answer} 2: Rila. \textbf{Type}: Unpopular\\
\hline
\textbf{Triplets:} [[Nicu Stroia, participant in, 1992 Summer Olympics], [1992 Summer Olympics, country, Spain], [Spain, capital, Madrid]]\\
\textbf{Main Question:} What is the capital of the country where Nicu Stroia participated in an event?\\
\textbf{Main Answer:} Madrid\\
\textbf{Subquestion pairs:}\\
\textbf{Sub-question} 0: In which events or activities did Nicu Stroia participate? \textbf{Sub-answer} 0: 1992 Summer Olympics. \textbf{Type}: Unpopular\\
\textbf{Sub-question} 1: In which country were the 1992 Summer Olympics held? \textbf{Sub-answer} 1: Spain. \textbf{Type}: Popular\\
\textbf{Sub-question} 2: What is the capital of Spain? \textbf{Sub-answer} 2: Madrid. \textbf{Type}: Popular\\
\hline
\textbf{Triplets:} [[Bunk Moreland, present in work, The Wire], [The Wire, original broadcaster, HBO], [HBO, parent organization, WarnerMedia]]\\
\textbf{Main Question:} What is the parent organization of the original broadcaster of the work featuring Bunk Moreland?\\
\textbf{Main Answer:} WarnerMedia\\
\textbf{Subquestion pairs:}\\
\textbf{Sub-question} 0: In which work does the character Bunk Moreland appear? \textbf{Sub-answer} 0: The Wire. \textbf{Type}: Unpopular\\
\textbf{Sub-question} 1: What is the original broadcaster of The Wire? \textbf{Sub-answer} 1: HBO. \textbf{Type}: Popular\\
\textbf{Sub-question} 2: What is the parent organization of HBO? \textbf{Sub-answer} 2: WarnerMedia. \textbf{Type}: Unpopular\\
\hline
\textbf{Triplets:} [[Ewout van Asbeck, sport, field hockey], [field hockey, country of origin, England], [England, capital, London]]\\
\textbf{Main Question:} What is the capital of the country of origin of the sport in which Ewout van Asbeck participates?\\
\textbf{Main Answer:} London\\
\textbf{Subquestion pairs:}\\
\textbf{Sub-question} 0: What sport does Ewout van Asbeck participate in? \textbf{Sub-answer} 0: field hockey. \textbf{Type}: Unpopular\\
\textbf{Sub-question} 1: Which country is the origin of field hockey? \textbf{Sub-answer} 1: England. \textbf{Type}: Unpopular\\
\textbf{Sub-question} 2: What is the capital of England? \textbf{Sub-answer} 2: London. \textbf{Type}: Popular\\
\hline
\textbf{Triplets:} [[College Hockey in the D, sport, ice hockey], [ice hockey, authority, International Ice Hockey Federation], [International Ice Hockey Federation, headquarters location, Zürich]]\\
\textbf{Main Question:} Where is the headquarters of the authority governing the sport of College Hockey in the D located?\\
\textbf{Main Answer:} Zürich\\
\textbf{Subquestion pairs:}\\
\textbf{Sub-question} 0: What sport is associated with College Hockey in the D? \textbf{Sub-answer} 0: ice hockey. \textbf{Type}: Unpopular\\
\textbf{Sub-question} 1: Which organization is the governing authority for ice hockey? \textbf{Sub-answer} 1: International Ice Hockey Federation. \textbf{Type}: Unpopular\\
\textbf{Sub-question} 2: Where are the headquarters of the International Ice Hockey Federation located? \textbf{Sub-answer} 2: Zürich. \textbf{Type}: Unpopular\\
\hline
\end{tabular}\caption{Three-hop question-answer pairs and their corresponding types in \textsc{MINTQA-pop}.}\end{table*}
\begin{table*}[ht]\centering\scriptsize\begin{tabular}{p{15cm}}\hline
\textbf{Triplets:} [[National Hockey League, sport, ice hockey], [ice hockey, authority, International Ice Hockey Federation], [International Ice Hockey Federation, country, Switzerland], [Switzerland, continent, Europe]]\\
\textbf{Main Question:} On which continent is the country that has authority over the sport played in the National Hockey League located?\\
\textbf{Main Answer:} Europe\\
\textbf{Subquestion pairs:}\\
\textbf{Sub-question} 0: What sport is played in the National Hockey League? \textbf{Sub-answer} 0: ice hockey. \textbf{Type}: Popular\\
\textbf{Sub-question} 1: Which organization is the governing authority for ice hockey? \textbf{Sub-answer} 1: International Ice Hockey Federation. \textbf{Type}: Unpopular\\
\textbf{Sub-question} 2: Which country is the International Ice Hockey Federation based in? \textbf{Sub-answer} 2: Switzerland. \textbf{Type}: Unpopular\\
\textbf{Sub-question} 3: On which continent is Switzerland located? \textbf{Sub-answer} 3: Europe. \textbf{Type}: Unpopular\\
\hline
\textbf{Triplets:} [[Rafael Bejarano, place of birth, Arequipa], [Arequipa, country, Peru], [Peru, capital, Lima], [Lima, located in or next to body of water, Rímac River]]\\
\textbf{Main Question:} Which body of water is the capital of the country where Rafael Bejarano was born located next to?\\
\textbf{Main Answer:} Rímac River\\
\textbf{Subquestion pairs:}\\
\textbf{Sub-question} 0: Where was Rafael Bejarano born? \textbf{Sub-answer} 0: Arequipa. \textbf{Type}: Unpopular\\
\textbf{Sub-question} 1: In which country is Arequipa located? \textbf{Sub-answer} 1: Peru. \textbf{Type}: Popular\\
\textbf{Sub-question} 2: What is the capital of Peru? \textbf{Sub-answer} 2: Lima. \textbf{Type}: Popular\\
\textbf{Sub-question} 3: Which body of water is Lima located next to? \textbf{Sub-answer} 3: Rímac River. \textbf{Type}: Unpopular\\
\hline
\textbf{Triplets:} [[The Perfect Cocktail, part of the series, How I Met Your Mother], [How I Met Your Mother, original broadcaster, CBS], [CBS, owned by, Paramount Global], [Paramount Global, industry, mass media]]\\
\textbf{Main Question:} In which industry does the owner of the original broadcaster of the series that includes "The Perfect Cocktail" operate?\\
\textbf{Main Answer:} mass media\\
\textbf{Subquestion pairs:}\\
\textbf{Sub-question} 0: Of which series is "The Perfect Cocktail" a part? \textbf{Sub-answer} 0: How I Met Your Mother. \textbf{Type}: Unpopular\\
\textbf{Sub-question} 1: Which network originally broadcasted "How I Met Your Mother"? \textbf{Sub-answer} 1: CBS. \textbf{Type}: Popular\\
\textbf{Sub-question} 2: Who owns CBS? \textbf{Sub-answer} 2: Paramount Global. \textbf{Type}: Unpopular\\
\textbf{Sub-question} 3: In which industry does Paramount Global operate? \textbf{Sub-answer} 3: mass media. \textbf{Type}: Unpopular\\
\hline
\textbf{Triplets:} [[Saint George Killing the Dragon, creator, Bernat Martorell], [Bernat Martorell, place of death, Barcelona], [Barcelona, country, Spain], [Spain, capital, Madrid]]\\
\textbf{Main Question:} What is the capital of the country where the creator of Saint George Killing the Dragon died?\\
\textbf{Main Answer:} Madrid\\
\textbf{Subquestion pairs:}\\
\textbf{Sub-question} 0: Who is the creator of Saint George Killing the Dragon? \textbf{Sub-answer} 0: Bernat Martorell. \textbf{Type}: Unpopular\\
\textbf{Sub-question} 1: Where did Bernat Martorell die? \textbf{Sub-answer} 1: Barcelona. \textbf{Type}: Unpopular\\
\textbf{Sub-question} 2: In which country is Barcelona located? \textbf{Sub-answer} 2: Spain. \textbf{Type}: Popular\\
\textbf{Sub-question} 3: What is the capital of Spain? \textbf{Sub-answer} 3: Madrid. \textbf{Type}: Popular\\
\hline
\textbf{Triplets:} [[DWNX-FM, owned by, Radio Mindanao Network], [Radio Mindanao Network, headquarters location, Makati], [Makati, country, Philippines], [Philippines, continent, Asia]]\\
\textbf{Main Question:} On which continent is the country located where the headquarters of the owner of DWNX-FM is situated?\\
\textbf{Main Answer:} Asia\\
\textbf{Subquestion pairs:}\\
\textbf{Sub-question} 0: Who owns DWNX-FM? \textbf{Sub-answer} 0: Radio Mindanao Network. \textbf{Type}: Unpopular\\
\textbf{Sub-question} 1: Where is the headquarters of Radio Mindanao Network located? \textbf{Sub-answer} 1: Makati. \textbf{Type}: Unpopular\\
\textbf{Sub-question} 2: In which country is Makati located? \textbf{Sub-answer} 2: Philippines. \textbf{Type}: Popular\\
\textbf{Sub-question} 3: On which continent is the Philippines located? \textbf{Sub-answer} 3: Asia. \textbf{Type}: Unpopular\\
\hline
\textbf{Triplets:} [[2008 FA Trophy Final, location, Wembley Stadium], [Wembley Stadium, owned by, The Football Association], [The Football Association, applies to jurisdiction, England], [England, capital, London]]\\
\textbf{Main Question:} What is the capital of the jurisdiction that owns the location of the 2008 FA Trophy Final?\\
\textbf{Main Answer:} London\\
\textbf{Subquestion pairs:}\\
\textbf{Sub-question} 0: Where was the 2008 FA Trophy Final held? \textbf{Sub-answer} 0: Wembley Stadium. \textbf{Type}: Unpopular\\
\textbf{Sub-question} 1: Who owns Wembley Stadium? \textbf{Sub-answer} 1: The Football Association. \textbf{Type}: Unpopular\\
\textbf{Sub-question} 2: Which jurisdiction does The Football Association apply to? \textbf{Sub-answer} 2: England. \textbf{Type}: Unpopular\\
\textbf{Sub-question} 3: What is the capital of England? \textbf{Sub-answer} 3: London. \textbf{Type}: Popular\\
\hline
\textbf{Triplets:} [[Rothschild banking family of France, founded by, James Mayer de Rothschild], [James Mayer de Rothschild, place of birth, Frankfurt], [Frankfurt, located in or next to body of water, Main], [Main, mouth of the watercourse, Rhine]]\\
\textbf{Main Question:} Into which body of water does the river located next to the birthplace of the founder of the Rothschild banking family of France ultimately flow?\\
\textbf{Main Answer:} Rhine\\
\textbf{Subquestion pairs:}\\
\textbf{Sub-question} 0: Who founded the Rothschild banking family of France? \textbf{Sub-answer} 0: James Mayer de Rothschild. \textbf{Type}: Unpopular\\
\textbf{Sub-question} 1: Where was James Mayer de Rothschild born? \textbf{Sub-answer} 1: Frankfurt. \textbf{Type}: Unpopular\\
\textbf{Sub-question} 2: Which body of water is Frankfurt located next to? \textbf{Sub-answer} 2: Main. \textbf{Type}: Unpopular\\
\textbf{Sub-question} 3: Into which watercourse does the Main River flow? \textbf{Sub-answer} 3: Rhine. \textbf{Type}: Unpopular\\
\hline
\end{tabular}
\caption{Four-hop question-answer pairs and their corresponding types in \textsc{MINTQA-pop}.}
\label{last-case}
\end{table*}


\end{document}